\documentclass[10pt,twocolumn,letterpaper]{article}
\usepackage{wrapfig}
\usepackage{wacv}              

\definecolor{wacvblue}{rgb}{0.21,0.49,0.74}
\usepackage[pagebackref,breaklinks,colorlinks,allcolors=wacvblue]{hyperref}
\usepackage[normalem]{ulem}
\usepackage{mdframed}
\usepackage{multirow}
\usepackage{adjustbox}
\definecolor{lightgray}{gray}{0.95}
\newmdenv[
  backgroundcolor=lightgray,
  linewidth=0.5pt,
  roundcorner=5pt,
  skipabove=10pt,
  skipbelow=10pt,
  innertopmargin=5pt,
  innerbottommargin=5pt,
  innerleftmargin=10pt,
  innerrightmargin=10pt,
  tikzsetting={
    \tikz[overlay,remember picture] \node[anchor=south west, fill=black, opacity=0.15, xshift=5pt, yshift=-5pt] at (frame.south west) [minimum width=\textwidth, minimum height=\textheight] {};
  }
]{promptbox}
\usepackage{array}
\usepackage{xcolor}

\newcolumntype{G}{>{\color{blue}}c}

\newcolumntype{L}[1]{>{\raggedright\let\newline\\\arraybackslash\hspace{0pt}}m{#1}}
\newcolumntype{C}[1]{>{\centering\let\newline\\\arraybackslash\hspace{0pt}}m{#1}}
\newcolumntype{R}[1]{>{\raggedleft\let\newline\\\arraybackslash\hspace{0pt}}m{#1}}
 \expandafter\def\expandafter\normalsize\expandafter{%
    \normalsize
    \setlength\abovedisplayskip{3pt}
    \setlength\belowdisplayskip{2pt}
    \setlength\abovedisplayshortskip{1pt}
    \setlength\belowdisplayshortskip{1pt}
}

\usepackage{amsmath}
\makeatletter
\@namedef{ver@everyshi.sty}{}
\makeatother
\usepackage{tikz}
\usepackage{pifont}

\usepackage{amsmath,amssymb,amsfonts,amsthm}

\usepackage{lscape}

\newcommand{\fire}{%
    \begin{tikzpicture}[scale=0.2]
        \fill[red!80!yellow] (0,0) -- (0.2,0.5) -- (0,1) -- (-0.2,0.5) -- cycle; 
        \fill[orange] (0,-0.3) arc[start angle=-90,end angle=90,radius=0.3] -- cycle; 
    \end{tikzpicture}%
}

\newcommand{\cut}[1]{{}}
\newcommand{\ours}[1]{FLORA}

\def\wacvPaperID{105} 
\def\confName{WACV}
\def\confYear{2026}

\title{Dressing the Imagination: A Dataset for AI-Powered Translation of Text into Fashion Outfits and A Novel NeRA Adapter for Enhanced Feature Adaptation}

\author{Gayatri Deshmukh\\
Independent Researcher\\
{\tt\small dgayatri9850@gmail.com}
\and
Somsubhra De\\
IIT Madras\\
{\tt\small 22f3002680@ds.study.iitm.ac.in }
\and
Chirag Sehgal\\
Delhi Technological University\\
{\tt\small chiragsehgal224@gmail.com}
\and
Jishu Sen Gupta\\
IIT BHU\\
{\tt\small jishusen.gupta.mat22@itbhu.ac.in}
\and
Sparsh Mittal\\
IIT Roorkee\\
{\tt\small sparsh.mittal@ece.iitr.ac.in}
}
\begin{document}
\maketitle
\begin{abstract}
    Specialized datasets that capture the fashion industry's rich language and styling elements can boost progress in AI-driven fashion design. We present \textbf{FLORA} (\textbf{F}ashion \textbf{L}anguage \textbf{O}utfit \textbf{R}epresentation for \textbf{A}pparel Generation), the first comprehensive dataset containing 4,330 curated pairs of fashion outfits and corresponding textual descriptions. Each description utilizes industry-specific terminology and jargon commonly used by professional fashion designers, providing precise and detailed insights into the outfits. Hence, the dataset captures the delicate features and subtle stylistic elements necessary to create high-fidelity fashion designs. 
    We demonstrate that fine-tuning generative models on the FLORA dataset significantly enhances their capability to generate accurate and stylistically rich images from textual descriptions of fashion sketches. FLORA will catalyze the creation of advanced AI models capable of comprehending and producing subtle, stylistically rich fashion designs.  It will also help fashion designers and end-users to bring their ideas to life.   
    As a second orthogonal contribution, we introduce \textbf{NeRA} (\textbf{N}onlinear low-rank \textbf{E}xpressive \textbf{R}epresentation \textbf{A}dapter), a novel adapter architecture based on Kolmogorov-Arnold Networks (KAN). Unlike traditional PEFT techniques such as LoRA, LoKR, DoRA, and LoHA that use MLP adapters, NeRA uses learnable spline-based nonlinear transformations, enabling superior modeling of complex semantic relationships, achieving strong fidelity, faster convergence and semantic alignment. Extensive experiments  on our proposed FLORA and LAION-5B datasets validate the superiority of NeRA over existing adapters. 
    See \href{https://candlelabai.github.io/WACV2026-FLORA-Dataset-NeRA-Adapter/}{Project page} for data and code\footnote{This work is supported by the CRG/2022/003821 project of Science and Engineering Research Board (SERB) of India. Gayatri, Somsubhra, Chirag and Jishu contributed to this project while working as interns at IIT Roorkee.}.
\end{abstract}

 \section{Introduction}
\label{sec:introduction_Section}
Artificial intelligence has become increasingly influential in various industries, fashion being a prominent example. The global fashion industry, valued at $2.4$ trillion, has been growing at an annual rate of $5.5\%$ over the past decade \cite{mckinsey2022stateoffashion}.
Major fashion brands  are leveraging AI to enhance creativity,  efficiency and consumer personalization \cite{fashionai1, fashionai2}. 
\cut{AI tools now support designers in generating patterns and styles by analyzing consumer preferences, allowing for faster, trend-responsive design cycles.} 
AI's role in fashion ranges from virtual try-on systems to trend forecasting from social media and fashion show datasets, product development, supply chain optimization, and sustainable practices \cite{sustfashion}. 

Despite the potential, AI applications in fashion face several challenges \cite{8601258}. One major issue is the complexity of fashion design, which involves subjective details like color, style, fabric, and texture. AI models often struggle to capture the nuanced details of fashion sketches. Existing models often produce outputs that lack coherence or fail to align with the provided descriptions \cite{surveytext2img,KhushbooPatel2024,foong2023challengesimagegenerationmodels}. Another challenge is the unavailability of well-annotated datasets, especially sketch-based datasets. Traditional fashion design practices, including seasonal collections and lengthy production cycles, often lag behind the fast-paced demands of consumers. As trends evolve rapidly, designers face the challenge of producing timely collections that resonate with audiences. This underscores the need for innovations that streamline design processes to maintain competitiveness. 

This research addresses the problem of generating high-fidelity fashion outfit sketches based on detailed text descriptions. Unlike virtual try-on systems \cite{10213129}, which simulate how garments fit on a user, our goal is to create accurate and visually appealing fashion sketches from textual information. Our paper addresses the critical issue of dataset quality by providing a meticulously curated and annotated dataset. Developing a robust pipeline for sketch generation from text can help overcome obstacles in translating descriptive text into coherent and stylistically appropriate visual representations.
With this, we aim to advance the capability of AI in fashion design to streamline the design workflow, reducing the time spent on initial sketches, enhancing collaboration, and boosting sales. This will help emerging designers who may not have traditional sketching skills and can serve as a valuable tool for designers, researchers, and other stakeholders in this industry.

Additionally, in today’s landscape, datasets are increasingly complex, with higher levels of non-linearity that make it challenging for models to effectively capture intricate patterns. Models often struggle to adapt to these complexities. For instance, datasets like FLORA present unique challenges due to their emphasis on fine-grained details such as fabric type, fit, and decorative elements which require nuanced understanding and multi-modal alignment. These subtle attributes introduce high variability and complexity, making it difficult for models to generalize well. 
To address this, we propose NeRA (Nonlinear low-rank Expressive Representation Adapter), a new adapter architecture designed to capture fine-grained, nonlinear dependencies in complex datasets like FLORA. NeRA leverages spline-based activations inspired by Kolmogorov–Arnold Networks to enable expressive and flexible feature transformation, improving ability to capture complex, fine-grained relationships. We  make the following key contributions:
 
\begin{itemize}\itemsep0pt
        \item \textbf{FLORA Dataset:} We introduce FLORA, a large-scale, curated dataset comprising 4,330 pairs of fashion outfit sketches and detailed textual descriptions. Each description uses industry-specific terminology, capturing the nuanced design elements essential for high-fidelity fashion sketch generation from text. FLORA fills a critical gap in existing resources by providing the first dataset tailored specifically to text-to-fashion sketch generation. FLORA will allow AI models to create precise, high-quality fashion sketches from text, easing reliance on manual sketching and accelerating early design phases.
        
        \item \textbf{NeRA Adapter:} 
        We propose NeRA, a nonlinear low-rank adapter architecture based on Kolmogorov-Arnold Networks (KANs)~\cite{liu2024kankolmogorovarnoldnetworks}.  By leveraging learnable spline-based activations, NeRA enables more expressive and flexible feature transformations than MLP-based adapters such as LoRA, LoKR, DoRA, and LoHA, improving semantic alignment and fidelity in generative tasks. Its parameter efficiency allows large diffusion models to adapt to fine-grained tasks like fashion sketch synthesis.

        \item Experiments show that NeRA achieves faster convergence and higher fidelity than MLP based adapters, enabling models to reach optimal performance in fewer training steps. Additionally, fine-tuning generative models on the FLORA dataset greatly improves their ability to produce accurate, stylistically rich images from fashion outfit descriptions.
    \end{itemize}

    \begin{figure*}
\centering  
  \includegraphics[width=.55\linewidth]{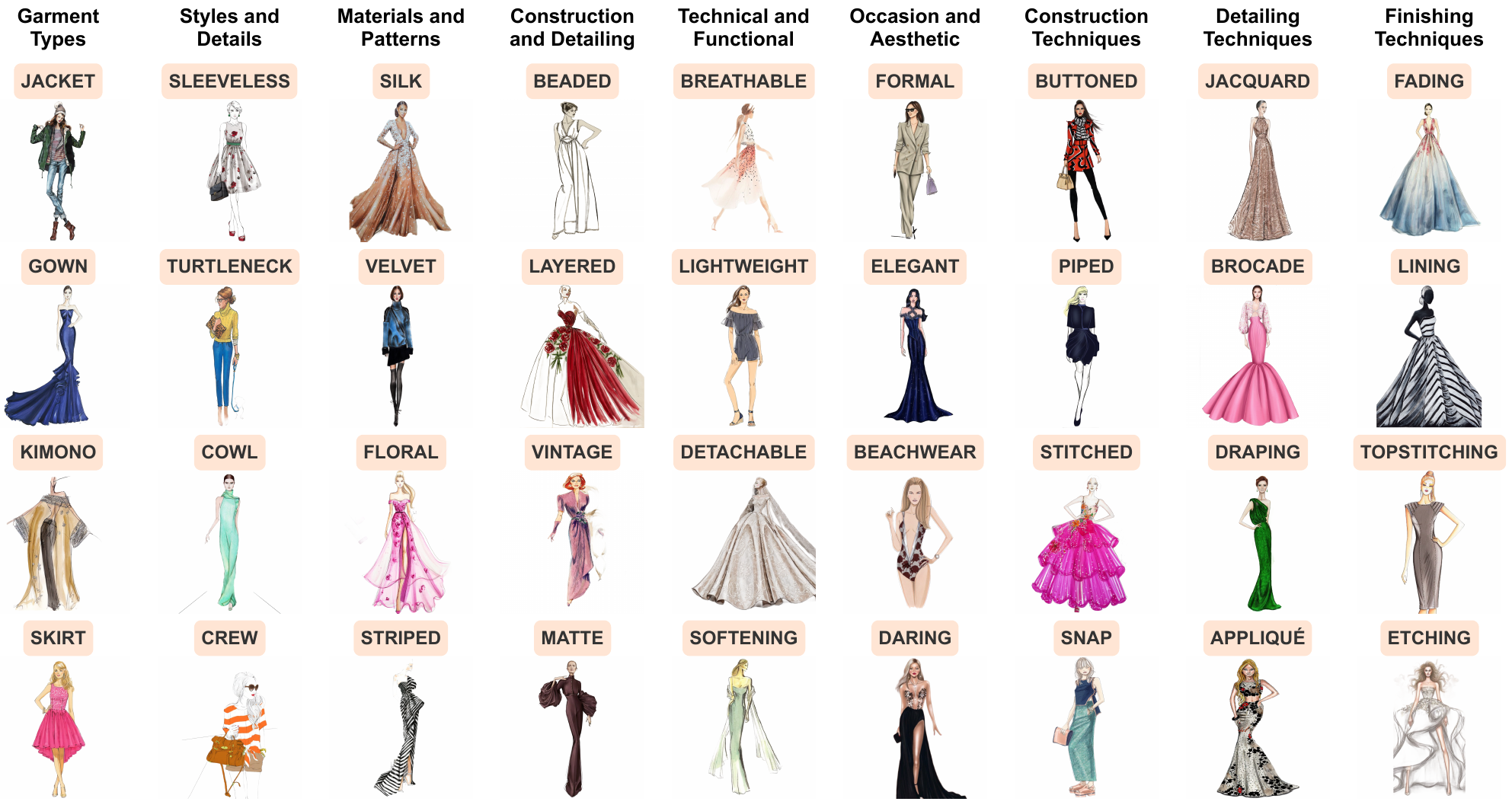}
  \hspace{15pt} 
  \includegraphics[scale=0.3]{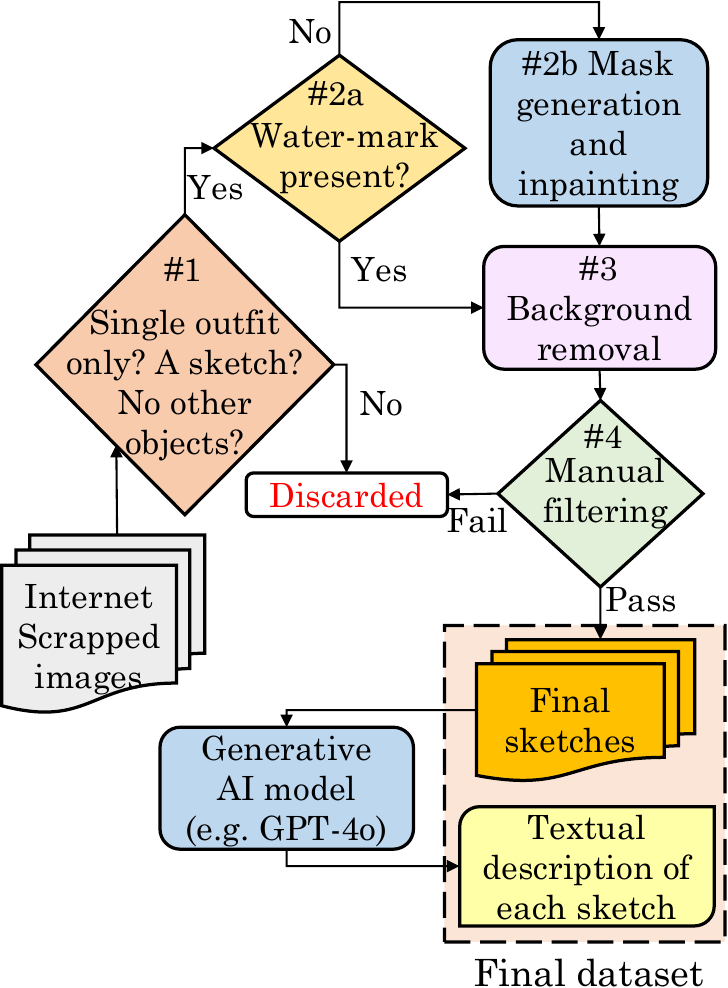}
  \caption{Left: Selected classes from each of the 9 categories in FLORA dataset, showcasing its diversity  (refer Supplementary for details). Right: Image filtering steps for creating the training dataset}
  \label{fig:dataset}
\end{figure*}

\section{Related Works}
\label{sec:related_work}
\textbf{Related work on AI in fashion:} 
Style transfer models, e.g., Pix2Pix \cite{pix2pix2017}, use conditional GANs \cite{mirza2014conditionalgenerativeadversarialnets} for image translation tasks. FashionGAN \cite{fashionGAN} allows designers to create virtual model visualizations of new apparel designs.
Recently, diffusion models have surpassed GANs in  image generation  \cite{10.1145/3600211.3604681}. 
Transformer based text-to-image models, such as DALL-E \cite{ramesh2021zeroshottexttoimagegeneration}, Imagen \cite{saharia2022photorealistictexttoimagediffusionmodels}, and Stable Diffusion \cite{Stable-Diffusion-1.5, Stable-Diffusion-3} now enable AI to create garment visuals from descriptions.

However, challenges persist in generating high-fidelity and stylistically accurate images from complex fashion descriptions, particularly for stylized sketches, which are integral to fashion design. While these models excel at realistic image generation, they lack optimization for fashion sketches. Further, there are no dedicated datasets that are sufficiently curated and explicitly annotated for generating fashion sketches from text descriptions. This limits the development of robust models for fashion sketch generation.

Recent datasets like Market-1501 \cite{Zheng_2015_ICCV}, DeepFashion \cite{liuLQWTcvpr16DeepFashion}, and VITON \cite{han2017viton} support fashion synthesis applications such as human parsing, pose estimation, landmark detection, style transfer, and clothing simulation.
DeepFashion and DeepFashion2 \cite{ge2019deepfashion2} have enabled tasks such as garment recognition, landmark detection, and segmentation by providing high-quality annotations of real-world fashion images. FashionGen \cite{rostamzadeh2018fashion} contributed richly annotated catalog images with class labels suitable for GAN-based generation, while DeepFashion-Multimodal \cite{jiang2022text2human} further incorporated textual descriptions to support cross-modal retrieval and generation. VITON-HD \cite{choi2021viton} has been pivotal for virtual try-on tasks, offering paired images of individuals in different outfits. \cite{baldrati2023multimodal} provide higher resolution try-on datasets with multiple clothing categories and annotations. However, these too are designed for virtual try-on synthesis and contain studio-quality model images rather than conceptual design sketches. Likewise, the Fashion Synthesis dataset introduced in the same work supports latent diffusion-based outfit generation but still remains grounded in posed model photography rather than design-centric imagery. 
Despite their utility, all of these datasets rely on photographic images and lack the abstraction and creative intent represented in designer sketches. \emph{To the best of our knowledge, no existing dataset includes detailed, text-paired fashion sketches an essential modality in the early phases of fashion design.}
We address this gap by proposing a novel dataset tailored for generating fashion sketches from textual descriptions. FLORA is the first such dataset, comprising high-quality sketch-text pairs with rich, domain-specific descriptions that reflect the professional vocabulary and compositional detail found in designer workflows. 
\cut{This focus lays a foundation for advanced AI applications in fashion design.}

 \textbf{Related work on Parameter Efficient Fine-Tuning (PEFT) Method:}
Low-Rank Adaptation (LoRA) \cite{hu2021lora} enables efficient fine-tuning of large models. Instead of updating all model parameters, LoRA freezes the original model weights and introduces a pair of trainable low-rank matrices. 
\cut{This significantly reduces the number of trainable parameters while maintaining model performance.} Formally, for a weight matrix \( W \in \mathbb{R}^{d \times d} \), the update \( \Delta W \) is expressed as:
\begin{equation} \Delta W = W_{\text{up}} W_{\text{down}} \end{equation}
where \( W_{\text{up}} \in \mathbb{R}^{d \times l} \) and \( W_{\text{down}} \in \mathbb{R}^{l \times d} \), with \( l \ll d \), minimizing computational and memory overhead. DoRA \cite{liu2024dora} improves flexibility by decomposing weights into magnitude and direction, enhancing convergence. Variants, such as LoHa \cite{hyeon2021fedpara} and LoKr \cite{yeh2023navigating}, introduce element-wise and Kronecker product adaptations for more expressive yet efficient tuning. However, these methods rely mostly on linear or shallow nonlinear operations, limiting their ability to capture complex, nonlinear relationships critical for detailed, domain-specific tasks. These approaches struggle to fully model the rich, nonlinear dependencies required in creative fields such as fashion design. Our proposed NeRA enhances adapter expressiveness and adaptability for such complex modeling challenges.

\textbf{Related work on KANs:}
Kolmogorov-Arnold Networks (KANs) \cite{liu2024kankolmogorovarnoldnetworks}, inspired by the Kolmogorov-Arnold representation theorem, offer an interpretable and flexible alternative to traditional MLPs. With learnable edge activations, KANs adaptively learn input relationships, making them effective for complex function approximation and interpretability in data patterns.
Fast KAN \cite{li2024u} introduced an adaptation where B-splines are replaced by Radial Basis Functions (RBFs), aiming to reduce the computational overhead associated with splines. ReLU-KAN \cite{qiu2024relu} addresses the computational complexity in traditional KANs by using ReLU and point-wise multiplication, combining KANs' ``catastrophic forgetting avoidance" with ReLU's efficiency, making it suitable for inference and training in deep learning frameworks.
ConvKANs \cite{bodner2024convolutional} integrate KAN nonlinear activations into convolutional layers, making them valuable for interpretable applications in computer vision. U-KAN  \cite{li2024u} integrates KAN layers into U-Net for efficient medical image segmentation. Kolmogorov-Arnold Transformer (KAT) \cite{yang2024kolmogorov} outperforms ViT models on vision tasks by incorporating KANs into MLP-based transformers.

\section{Proposed FLORA Dataset}
\label{sec:dataset_Section}
We introduce FLORA, a large-scale dataset comprising 4,330 outfit sketch and textual description pairs. \cref{fig:dataset} (left) shows representative categories of image (more details in supplementary) from the dataset. This dataset aims to facilitate the training of generative models for fashion image synthesis, designed specifically for generating fashion sketches based on descriptive inputs. It represents the first dataset of its kind, addressing the significant gap in resources available for sketch-based fashion design.

\textbf{Collection of dataset:}
To create a diverse range of fashion-related sketches, we initiated the process by web-scraping images using various search queries such as `fashion-outfit sketches', `fashion illustrations', `wedding-gown sketches' and `fashion pencil sketches'. We, thus, collected 10,042 images. Since these images had various forms of noise, e.g., signatures, watermarks and text overlays, we employed a multi-stage filtering approach (right of Figure \ref{fig:dataset}) to obtain a suitable dataset.

\subsection{Multi-Stage Filtering of Images}
\begin{figure}[ht]
          \centering
          \includegraphics[width=0.4\textwidth]{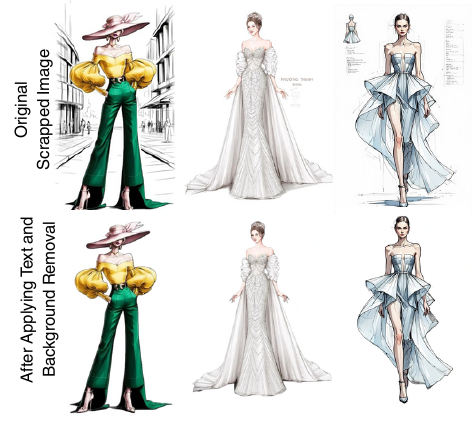}
          \caption{ Visual comparison of original and processed images.  } 
          \label{fig:text_and_bg_remove}
        \end{figure}

\textbf{1. Prompt-Based Filtering} We first implemented prompt-based filtering using LLaVa 32b \cite{liu2023llava}, a VLLM (Vision Large Language Model) model. It checks whether the image is a sketch with only a single outfit and no other objects.\\
\textbf{2. Watermark Removal} Removing signatures and text is crucial for maintaining images' visual integrity and improving the model's ability to learn and generalize effectively. For images with watermark, we applied Keras-OCR \cite{kerasocr} for text detection and mask generation followed by `ControlNet Stable Diffusion Scene-Text Eraser' \cite{scenetexteraser} for inpainting. This pipeline removed extraneous elements while preserving the original outfit details. Notably, this step was performed only to reduce visual distractions that could affect model training, and not to alter ownership attribution.\\
\textbf{3. Background Removal} We eliminated extraneous objects and color variations for images with distracting background noise. Specifically, we replaced the original backgrounds with clean, transparent ones using a background-remover model \cite{transparent-background}. This enhances visual clarity and ensures the focus remains on the main subject.  Figure \ref{fig:text_and_bg_remove} shows that the pipeline successfully removes extraneous text and backgrounds while preserving the integrity of the main object’s shape and fine texture details. Clearly, our preprocessing does not introduce artifacts or distortions that could negatively affect downstream model training.

\textbf{4. Manual Filtering} Finally, we manually checked each image to remove duplicates. We split the images containing multiple sketches into separate sketches to maximize the number of usable samples. We concluded with 4,330 clean and relevant images.

\subsection{Generating description of each sketch} 
\begin{figure}[htbp]
  \centering
  \includegraphics[width=.95\linewidth]{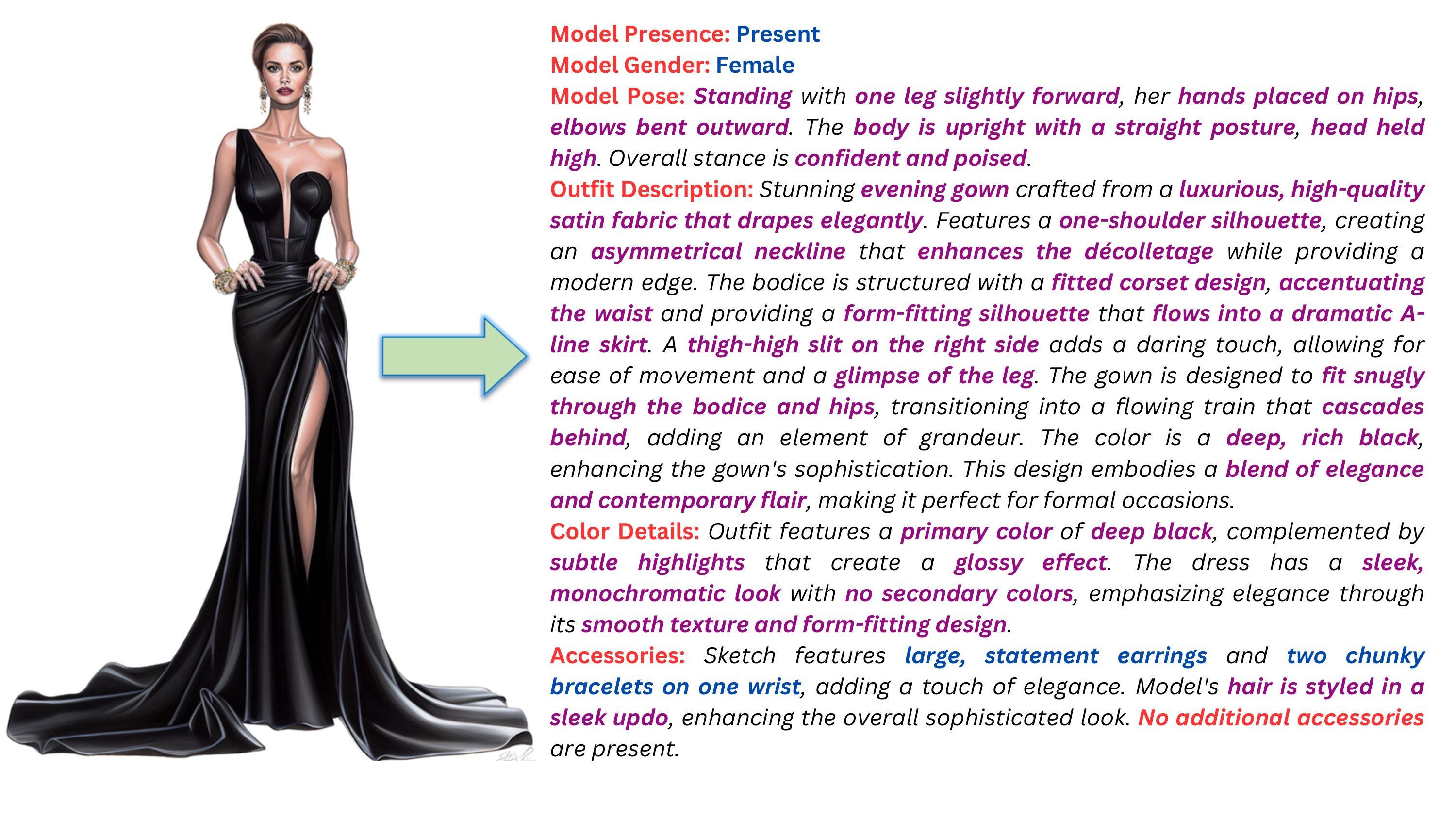}
  \captionof{figure}{A figure and the generated textual description pair}
  \label{fig:fashionoutfit}
\end{figure}

We utilized a multimodal model, viz., OpenAI's GPT-4o \cite{gpt4o} to generate detailed outfit descriptions from fashion sketches. Each outfit description is structured to include the \textit{human model pose} that consists of the model's posture, including the position of hands and legs, allowing for accurate sketch recreation, \textit{outfit details}, its \textit{color} and \textit{accessories} that lists any additional elements like shoes or jewelry.   \cref{fig:fashionoutfit}  shows an example of one such outfit description. 

To assess whether general-purpose models like GPT-4o are suitable for fashion description tasks, we compared them with domain-specific vision-language models such as FAME-ViL~\cite{han2023fame} and UniFashion~\cite{zhao2024unifashion}. The full experimental setup and results are provided in Section \ref{sec:ablation_study_Section}.  
Further details on the dataset and prompt design are also included in the supplementary section.

\section{Proposed NeRA Adapter}
As a second orthogonal contribution, we propose NeRA (Nonlinear low-rank Expressive Representation Adapter). The need for adaptable and computationally efficient models has grown recently, especially in domains where fine-tuning large models for downstream tasks can be computationally expensive. Traditional methods like LoRA modules address this need by introducing learnable parameters in a lower-dimensional space, typically employing MLPs for transformation. However, MLPs face limitations in efficiently modeling complex non-linear transformations. Our proposed NeRA overcomes these limitations. It leverages KANs with learnable spline-based activation functions. This architecture efficiently approximates complex, highly non-linear functions, enhancing parameter efficiency and flexibility. Replacing MLP-based LoRA with NeRA improves computational efficiency and adaptability, making it well-suited for tasks requiring sophisticated non-linear representations.    
We first provide an overview on KAN modules, and then discuss our proposed NeRA adapter.

\textbf{Background on KANs:} 
    KANs \cite{liu2024kankolmogorovarnoldnetworks} replace traditional MLP activations with learnable B-spline functions, offering greater flexibility and expressiveness with fewer parameters. B-splines allow localized adjustments while preserving smoothness and differentiability for efficient backpropagation. Based on the Kolmogorov-Arnold representation theorem, any high-dimensional function can be expressed as a sum of univariate functions.  
    In the KAN layer \cite{liu2024kankolmogorovarnoldnetworks}, activation is defined as:
    \begin{equation} 
    \varphi(x) = w \left( b(x) + \text{spline}(x) \right) 
    \end{equation}
    Here, $b(x) = \text{silu}(x) = {x}/{(1 + e^{-x})}$ and $spline(x)$ is a learnable linear combination of B-splines, enabling KANs to effectively model complex functions.

\begin{figure}[htbp]
        \centering       
        \includegraphics[width=0.6\columnwidth]{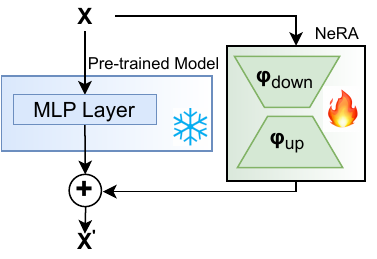}
        \caption{Integration of the NeRA with a pre-trained model. \textcolor{blue}{\ding{105}} and \(\fire\) indicate frozen and  trainable weights, respectively. }
        \label{fig:adapter}
   \end{figure}

\textbf{Proposed NeRA Adapter:} NeRA architecture (Figure \ref{fig:adapter}) incorporates KAN as an adapter layer alongside the pre-trained model's MLP layer. Similar to LoRA, these adapters can be added to any layer. The NeRA adapter layer processes input $X$, which includes down-projection ($\phi_{down}$) and up-projection ($\phi_{up}$) functions. These functions transform the input to a lower-dimensional space and then project it back, producing $X'$ using learnable, spline-based activation functions. 
Then, MLP and NeRA adapter outputs are aggregated and passed to the subsequent layer.

    As depicted in Figure \ref{fig:adapter}, the input \( X\in\{ x_1, x_2, x_3, \dots, x_m \}\)  is projected into a lower-dimensional representation \(H\in\{ h_1, h_2, h_3, \dots, h_n\}\), where \( X\in \mathbb{R}^m\), \( H\in \mathbb{R}^n\), with $n$ = $m/2$, through
    \begin{equation}
        \phi_{\text{down}}(X) = W_b^{\text{down}} \cdot g(X) + W_s^{\text{down}} \left( \sum\nolimits_{i=1}^{n} c_i \, \gamma(X) \right)
    \end{equation}
Here, \( W_b^{\text{down}} \) and \( W_s^{\text{down}} \) are learnable stabilizing matrices, \( g(X) \) is an activation function, and \( \gamma_i(X) \) represents a set of basis functions (such as B-splines or Radial Basis Functions). The terms \( c_i \) are the coefficients learned during training. The transformed representation \( H \) is projected back to the original feature space \( \mathbb{R}^m \).
    \begin{equation}
        \phi_{\text{up}}(X) = W_b^{\text{up}} \cdot g(H) + W_s^{\text{up}} \left( \sum\nolimits_{i=1}^{m} d_i \, \gamma(H) \right)
    \end{equation}
\( W_b^{\text{up}} \) and \( W_s^{\text{up}} \) are additional stabilization matrices, \( d_i \) are learned coefficients, and \( \gamma_i(H) \) is the basis function representation in the compressed space. This up-projection allows the model to re-expand the representation to the original space while retaining the most relevant features. Finally, outputs from the NeRA and MLP are aggregated as
    \begin{equation}
        X' = \mathcal{L}_{\theta}(X) + \left( \phi_{\text{up}} \cdot \phi_{\text{down}} \right)(X)
    \end{equation}

$\mathcal{L}_{\theta}(.)$ is the original pre-trained model. The final aggregated output $X'$ is then passed to subsequent layers, ensuring that the NeRA's non-linear transformations complement the MLP's capabilities. This complements the main model by introducing non-linear transformations, which traditional PEFT methods like LoRA’s MLP layers struggle to capture efficiently.

\begin{figure}[b]
        \centering
        \includegraphics[width=\columnwidth]{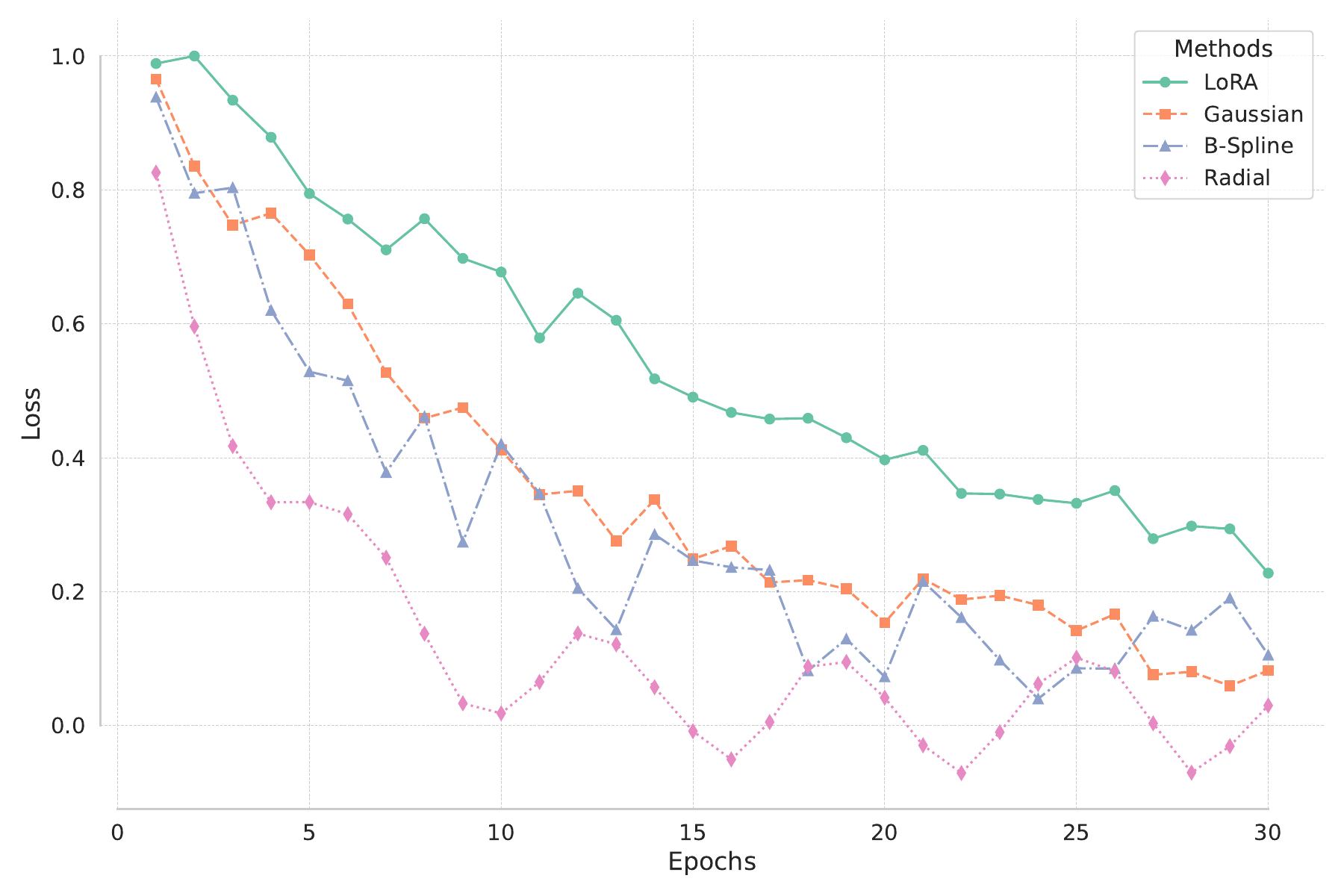}
        \caption{Comparison of LoRA and NeRA (using Gaussian, B-Spline, and Radial basis functions) on the FLUX model.}
        \label{fig:model_loss_curve}
\end{figure}

\textbf{Salient features:} NeRA differs fundamentally from LoRA's fixed MLP-based transformations by utilizing flexible, learnable basis functions. This enables NeRA to capture fine-grained patterns within the data, which is crucial for tasks where complex dependencies must be learned. While LoRA adapters are designed to prevent catastrophic forgetting by keeping the core model weights frozen, NeRA offer an additional layer of resilience through their unique use of adaptive, learnable activation functions. Unlike static activations in MLP-based LoRA, NeRA's non-linear functions can dynamically adjust to new tasks, selectively retaining important representations from prior tasks without overwriting them. This flexibility enables NeRA to handle the complexities of sequential learning better, as they can modulate responses specifically for new data while preserving critical knowledge from earlier tasks. This added adaptability has been shown (e.g., ReLU-KAN \cite{qiu2024relu}) to provide enhanced memory retention, making NeRA particularly suited for continual learning scenarios. Additionally, NeRA demonstrates faster convergence compared to LoRA (refer to Figure \ref{fig:model_loss_curve}), i.e., they can reach optimal or near-optimal performance in fewer training steps. \textit{We prove that NeRA is more expressive than LoRA (refer  supplementary).}

\section{Experimental Section}\label{sec:experimental_Section}

Detailed information regarding model training procedures and a comprehensive description of the FLORA test set creation process are provided in the supplementary material.

\setlength{\tabcolsep}{2pt}
\begin{table*}[ht]
\centering\footnotesize
\begin{tabular}{ccccccccccccc}
\hline
             & \multicolumn{2}{c}{Zero Shot} & \multicolumn{2}{c}{LoHA} & \multicolumn{2}{c}{DoRA} & \multicolumn{2}{c}{LoKR} & \multicolumn{2}{c}{LoRA} & \multicolumn{2}{c}{NeRA}          \\
Methods      & FID($\downarrow$)         & CLIPSIM($\uparrow$)      & FID($\downarrow$)      & CLIPSIM($\uparrow$)    & FID($\downarrow$)      & CLIPSIM($\uparrow$)    & FID($\downarrow$)      & CLIPSIM($\uparrow$)    & FID($\downarrow$)      & CLIPSIM($\uparrow$)    & FID($\downarrow$)            & CLIPSIM($\uparrow$)      \\ \hline
FLUX \cite{fluxdev}         & 09.09        & 0.3112          & 09.34     & 0.3001        & 09.01     & 0.3131        & 08.33     & 0.3223        & 07.88     & 0.2987        & \textbf{06.05}  & \textbf{0.3412} \\
SD \cite{Stable-Diffusion-1.5}         & 19.25       & 0.0623          & 20.01    & 0.0630         & 19.02    & 0.0720         & 19.33    & 0.0751        & 17.32    & 0.0910         & \textbf{16.51} & \textbf{0.1112} \\
SD-XL \cite{SD-XL}        & 16.37       & 0.1539          & 14.47    & 0.1667        & 15.89    & 0.1523        & 14.02    & 0.1782        & 15.32    & 0.1522        & \textbf{13.33} & \textbf{0.1897} \\
SD-3 \cite{Stable-Diffusion-3}         & 13.28       & 0.2118          & 13.62    & 0.2337        & 13.19    & 0.2279        & 13.01    & 0.2555        & 9.37     & 0.2119        & \textbf{08.07}  & \textbf{0.2503} \\
Pixart-Sigma \cite{pixart-sigma} & 14.11       & 0.2666          & 13.88    & 0.2424        & 13.88    & 0.2707        & 12.92    & 0.2992        & 09.89     & 0.2542        & \textbf{08.28}  & \textbf{0.2991} \\ \hline
\end{tabular}
\caption{Results of SOTA diffusion models for prompt-based outfit sketch generation using zero-shot, traditional MLP-based adapters, and proposed NeRA on \textbf{FLORA} dataset.}
\label{tab:results}
\end{table*}

\subsection{Quantitative Results}
\label{subsec:quantitative_results_Section}
We have conducted text-to-image generation experiments using several baseline models: Stable Diffusion (SD) \cite{Stable-Diffusion-1.5}, SD-XL \cite{SD-XL}, SD 3 \cite{Stable-Diffusion-3}, Pixart-Sigma \cite{pixart-sigma}, and FLUX \cite{fluxdev}. In the zero-shot setting, pre-trained baseline models are used to perform inference on our test set without any fine-tuning on FLORA. For NeRA, we used radial basis function. 
We evaluated each model's performance using FID (Fréchet Inception Distance) for image realism and CLIP-SIM for semantic alignment with textual prompts.

\begin{table}[!b]
\centering
 \resizebox{\columnwidth}{!}{
\begin{tabular}{lcccc}
\hline
\multicolumn{1}{c}{\multirow{2}{*}{Methods}}     & \multicolumn{2}{c}{LoRA} & \multicolumn{2}{c}{NeRA}          \\
\multicolumn{1}{c}{}                             & FID (↓)   & CLIPSIM (↑)  & FID (↓)        & CLIPSIM (↑)     \\ \hline
\multicolumn{5}{c}{Tested on LAION-5B Test Set}                                                                \\ \hline
FLUX\cite{fluxdev}              & 13.32     & 0.170       & \textbf{11.11} & \textbf{0.210} \\
SD\cite{Stable-Diffusion-1.5}   & 22.19     & 0.060       & \textbf{21.11} & \textbf{0.080} \\
SD-XL\cite{SD-XL}               & 21.01     & 0.070       & \textbf{20.93} & \textbf{0.101} \\
SD-3\cite{Stable-Diffusion-3}   & 18.88     & 0.110       & \textbf{17.99} & \textbf{0.140} \\
Pixart-Sigma\cite{pixart-sigma} & 19.02     & 0.130       & \textbf{17.76} & \textbf{0.170} \\ \hline
\multicolumn{5}{c}{Tested on FLORA Test Set}                                                                   \\ \hline
FLUX\cite{fluxdev}              & 20.78     & 0.100       & \textbf{19.96} & \textbf{0.150} \\
SD\cite{Stable-Diffusion-1.5}   & 33.17     & 0.002       & \textbf{32.07} & \textbf{0.010} \\
SD-XL\cite{SD-XL}               & 27.88     & 0.010       & \textbf{27.12} & \textbf{0.020} \\
SD-3\cite{Stable-Diffusion-3}   & 25.04     & 0.030       & \textbf{24.19} & \textbf{0.060} \\
Pixart-Sigma\cite{pixart-sigma} & 23.31     & 0.080       & \textbf{21.11} & \textbf{0.110} \\ \hline
\end{tabular}
}
\caption{Quantitative comparison on \textbf{LAION-5B} dataset. }
\label{tab:additional_dataset}
\end{table}

\textbf{Results on FLORA dataset:}
From Table \ref{tab:results}, the zero-shot results indicate that while these pre-trained models can produce realistic images to some extent, their semantic alignment with our specific fashion-focused textual prompts remains limited. The gap reflects FLORA's specialized fashion terms and novel design elements that pre-trained models struggle to capture.

We extended our experiments to compare NeRA with other state-of-the-art PEFT techniques such as LoRA, LoKR, DoRA, and LoHA. Across all base diffusion models, NeRA consistently outperforms all other adapters, demonstrating superior adaptability and expressive power. NeRA’s nonlinear spline-based design allows it to capture complex dependencies in a lower-dimensional yet flexible space, whereas other methods rely on primarily linear or shallow transformations, limiting their ability to model the nuanced relationships required for sketch generation. Among the base models, FLUX with NeRA achieved the best overall performance, with an FID of 6.05 and a CLIPSIM of 0.3412. Further, as shown in Figure \ref{fig:model_loss_curve}, NeRA demonstrates faster convergence and lower loss, indicating improved stability and performance in capturing relevant features.

\textbf{Results on LAION-5B dataset:}
To further validate NeRA's generalizability, we conducted additional experiments on the large-scale LIAON-5B \cite{schuhmann2022laion} dataset. It contains general-purpose image-text pairs. We fine-tuned models on LIAON-5B and evaluated them on both LIAON-5B and FLORA test sets. As shown in Table \ref{tab:additional_dataset}, NeRA remained effective across different domains, reinforcing its adaptability. However, baseline models fine-tuned on LIAON-5B exhibited a significant performance drop when evaluated on FLORA compared to those directly trained on FLORA (Table \ref{tab:results}). 
Evidently, domain-specific datasets are important for optimal performance. 
Despite this challenge, NeRA outperformed LoRA on all baselines, showing greater adaptability to complex, diverse feature distributions.

\begin{table}[htbp]
  \centering
  \resizebox{\columnwidth}{!}{
    \begin{tabular}{lcccc}\hline
          & Flux & SD-XL & SD-3 & Pixart \\
          & LoRA$|$NeRA & LoRA$|$NeRA & LoRA$|$NeRA & LoRA$|$ NeRA \\\hline
    Prompt Rel. & 59.78 $|$ 61.43 & 44.78 $|$ 47.12 & 53.32 $|$ 56.98 & 51.39 $|$ 55.03 \\
    Visual Quality & 45.25 $|$ 46.98 & 33.33 $|$ 36.54 & 40.14 $|$ 44.27 & 39.19 $|$ 41.14 \\
    Creativity & 52.29 $|$ 56.54 & 25.77 $|$ 29.30  & 37.82 $|$ 41.65 & 35.71 $|$ 40.21 \\
    Pose/Model acc & 65.19 $|$ 69.78 & 29.92 $|$ 33.47 & 59.12 $|$ 63.45 & 61.29 $|$ 65.43 \\
    Color Align. & 60.19 $|$ 63.33 & 48.93 $|$ 50.21 & 54.44 $|$ 57.02 & 53.53 $|$ 56.92 \\
    Contextual Fit & 45.92 $|$ 50.27 & 27.76 $|$ 31.19 & 39.31 $|$ 42.75 & 38.27 $|$ 40.91 \\\hline
    \end{tabular}}%
  \caption{Percentage of images with a score above 7.5 out of 10, as rated by 500 humans (acc.= accuracy, rel. = relevance, align. = alignment, SD results ommitted due to space limit).}
  \label{tab:human_eval}%
\end{table}%

\textbf{Human Evaluation Results:} We conducted a human evaluation on the FLORA test set, generating one sketch for each of 500 text descriptions. Each sketch was rated by a unique individual on parameters such as Prompt Relevance, Visual Quality, Creativity, Pose Accuracy, Color Alignment, Contextual Fit, and Overall Appeal (refer to supplementary for more details). As shown in Table \ref{tab:human_eval}, FLUX fine-tuned with NeRA achieved the highest scores, with NeRA consistently outperforming LoRA across all metrics.


    
\textbf{Efficiency, Inference Latency and Training Time:} Table \ref{tab:kan_vs_lora} compares the computational efficiency of each adapter across various baselines. 
NeRA generally has slightly higher inference times than LoRA. However, considering additional factors such as convergence speed during fine-tuning (Figure \ref{fig:model_loss_curve}) and NeRA’s effectiveness in adapting to non-linear patterns, this modest increase in inference time is a reasonable trade-off for the improved adaptability and performance that NeRA offers. Additionally, table compares parameter counts and training times for LoRA and NeRA. NeRA introduces a modest increase in parameters but significantly reduces training time; often by nearly half. 
    
\begin{table}[htbp]
    \centering
    \resizebox{\columnwidth}{!}{
    \begin{tabular}{c|cccccc}
    \hline
    Model           & \multicolumn{2}{c}{Latency (s)$\downarrow$} & \multicolumn{2}{c}{Parameters(M)$\downarrow$} & \multicolumn{2}{c}{Training Time (Hrs)$\downarrow$} \\ 
           & LoRA & NeRA           & LoRA & NeRA    & LoRA & NeRA \\ \midrule 
    FLUX   & 8.73 & 8.81        & 2.97 & 3.43          & 14.57 & 6.78               \\
    SD     & 13.43 & 14.42      & 1.47 & 1.82          & 9.37 & 5.62                \\
    SD-XL  & 9.87 & 10.02       & 2.81 & 3.65          & 9.23 & 4.97                \\
    SD-3   & 10.11 & 10.23      & 3.11 & 3.92          & 11.47 & 4.12               \\
    Pixart & 6.32 & 6.89        & 0.89 & 1.12          & 12.47 & 6.01               \\ \hline
    \end{tabular}
    }
    \caption{Latency, number of parameters and training time for LoRA and NeRA.  Number of diffusion steps were chosen to  achieve a target performance level: FLUX (20), SD (50), SD-XL and SD-3 (25), Pixart (15).}
    \label{tab:kan_vs_lora}
\end{table}

\begin{figure*}[htbp]
  \centering
   \includegraphics[width=\textwidth]{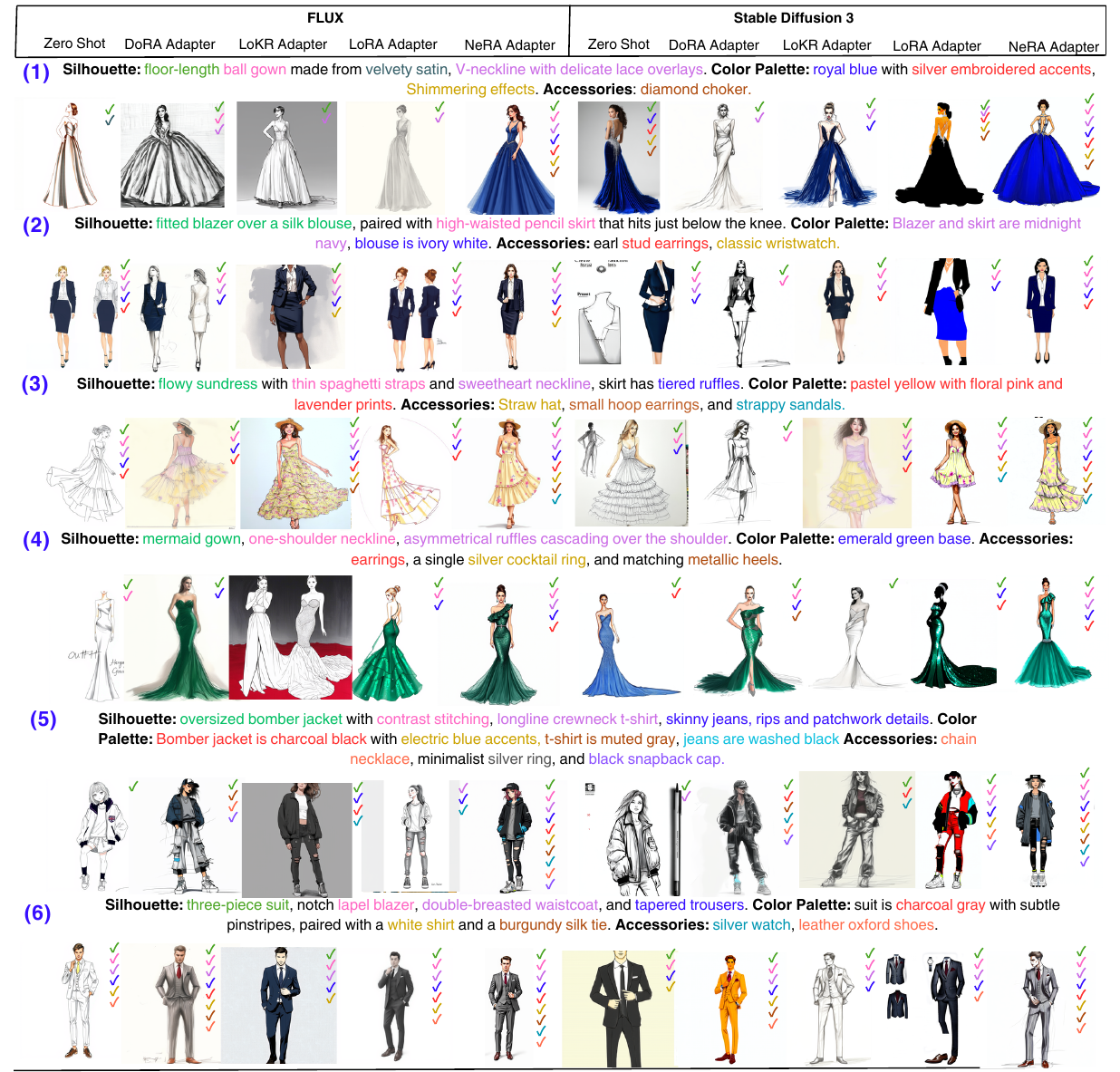}
   \caption{Qualitative comparison of zero-shot, MLP-based adapters and the proposed NeRA. Colored arrows mark whether the prompt elements written in that color were correctly generated.
   }
  \label{fig:outputcomparison}
\end{figure*}

\subsection{Qualitative Results}
\label{subsec:qualitative_results_Section}

Figure \ref{fig:outputcomparison} compares fashion sketch generation using zero-shot models, PEFT-based adapters (DoRA, LoKR, LoRA), and the proposed NeRA adapter with FLUX and Stable Diffusion 3. Zero-shot models often capture only basic silhouettes but fail to generate fine-grained fashion details like lace overlays, layered ruffles, precise colors, and accessories. For instance, in prompt (1), the zero-shot output shows only a plain gown without ornate details or jewelry, highlighting the limitations of relying solely on pre-trained weights for domain-specific tasks. 
 
PEFT-based adapters show incremental gains over zero-shot models. LoRA performs best among them in most of the cases, with stronger semantic alignment, especially for simpler outfits like the blazer and skirt in prompt (2). However, all three struggle with complex prompts. For instance in prompt (4), only  mermaid gown is generated, while one shoulder neckline, earrings are absent, revealing that their linear or shallow non-linear transformations cannot fully model the intricate relationships between garments, textures, and accessories in fashion design. 
 
The proposed NeRA adapter outperforms all other methods, showing superior fidelity and prompt adherence. Even in complex cases like prompt (5), featuring layered streetwear with patchwork textures and electric blue accents, NeRA accurately synthesizes the full look, including subtle details like a chain necklace and snapback. Its spline-based non-linear transformations enable richer semantic adaptation, capturing intricate relationships between silhouette, colors, and accessories.

\setlength{\tabcolsep}{2pt}

\begin{figure}[ht]
  \centering
  \includegraphics[width=0.45\textwidth]{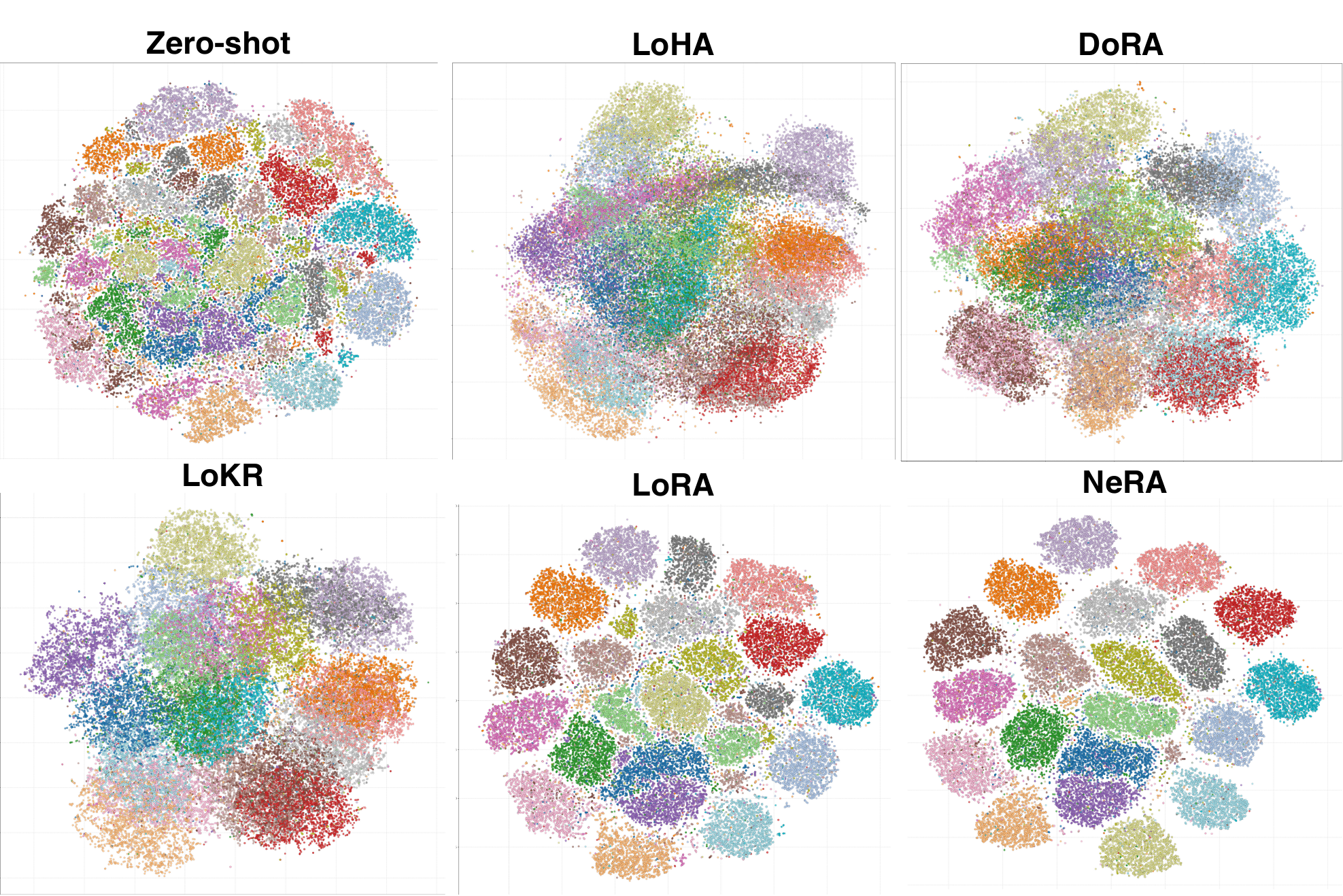}
  \caption{ t-SNE plots from the FLUX model for Zero-Shot, MLP-based, and NeRA outputs.    }
  \label{fig:tsne}
\end{figure}

Figure \ref{fig:tsne} presents the t-SNE visualization of the feature space by the Zero-Shot model and various adapters using the FLUX. Zero-shot outputs are scattered and overlapping, indicating weak semantic organization and poor disentanglement. The MLP-based adapters show partial clustering but still significant overlap, revealing their limited ability to model complex non-linear dependencies. NeRA produces well-separated, compact clusters, demonstrating its strength in structuring the latent space and learning fine-grained relationships between design components, leading to more accurate and stylistically coherent sketches. 

\subsection{Ablation Study}
\label{sec:ablation_study_Section}
We evaluate three types of multivariate function-approximating polynomials as basis functions in the NeRA adapter: B-Spline, RBF, and Gaussian.
As shown in Table \ref{tab:ablation}, the B-Spline basis was a middle-ground performer. 
B-splines, being piecewise polynomials, excel at providing localized control over the function, which is helpful for smooth approximations. However, unlike RBFs, they are sensitive to grid recalculations and scaling, reducing stability and efficiency in higher dimensions.

\setlength{\tabcolsep}{2pt}
\begin{table}[htbp]
\parbox{.4\linewidth}{
\centering
\footnotesize
\begin{tabular}{ccc}
\hline
NeRA Layers & FID & CS \\\midrule
B-Spline & 6.18 & 0.3002 \\
RBF & \textbf{6.09} & 0.3276 \\
Gaussian & 6.33 & \textbf{0.3301} \\
\hline
\end{tabular}
\caption{NeRA layer ablation results (CS=ClipSim) }\label{tab:ablation}
}
\hspace{5mm}
\parbox{.45\linewidth}{
\centering
\footnotesize
\begin{tabular}{ccccc}
\hline
& \multicolumn{2}{c}{LoRA} & \multicolumn{2}{c}{NeRA} \\\cmidrule(lr){2-3} \cmidrule(lr){4-5}
& FID & CS & FID & CS\\\hline
GPT-4o & \textbf{7.88} & \textbf{0.30} & \textbf{6.05} & \textbf{0.34} \\
FAME-ViL & 9.49 & 0.21 & 8.82 & 0.24 \\
UniFashion & 10.59 & 0.22 & 10.92 & 0.25 \\
\hline
\end{tabular}
\caption{FLUX performance on fashion specific models}\label{tab:flux-performance}
}
\end{table}

In contrast, the standard RBF delivered the lowest FID (6.09) and a strong CLIPSIM of 0.3276, achieving the best balance between sample realism and representational consistency. Unlike Gaussian RBFs, which rely on a fixed decay rate, standard RBFs are not bound to a specific decay, allowing them to adapt to complex patterns in data flexibly. This flexibility likely explains its superior FID, preserving realistic details better than Gaussian and B-Spline.

While slightly higher in FID (6.33), the Gaussian basis recorded the highest CLIPSIM at 0.3301, suggesting that it prioritizes structural consistency over fine-grained details. Gaussian functions have smooth and symmetrical decay properties, promoting stability in feature representation but at the cost of adaptability. This fixed decay makes it less effective at capturing localized detail than the general RBF, as seen in the higher FID score. Refer supplementary for an ablation study on the impact of NeRA downsample rates, LoRA rank and other finetuning techniques.

\textbf{Evaluating Description Quality Across Models:} To assess how the choice of vision-language model for generating outfit descriptions affects downstream performance, we conducted an ablation comparing three sources of descriptions: GPT-4o~\cite{gpt4o}, FAME-ViL~\cite{han2023fame}, and UniFashion~\cite{zhao2024unifashion}. For each model, we generated descriptions from fashion sketches to train a FLUX model, which was then evaluated on a test set. From Table \ref{tab:flux-performance}, FLUX trained on GPT-4o descriptions outperforms that trained on FAME-ViL or UniFashion descriptions in the NeRA setting.
\cut{As shown in Table \ref{tab:results}, FLUX trained on GPT-4o descriptions outperforms others in the NeRA setting, achieving a CLIPSIM of 0.3412 and FID of 6.05, compared to lower CLIPSIM (0.24 and 0.25) and higher FID (8.82 and 10.92) when trained on FAME-ViL or UniFashion descriptions (see Table \ref{tab:flux-performance}).} 
The performance gap stems not from the superiority of general-purpose models, but from differences in training data. Fashion-specific models like FAME-ViL and UniFashion are trained on structured, photorealistic images and catalog-style text, making them less effective on abstract, stylized sketches. They often miss sparse or symbolic cues and fine design details. In contrast, GPT-4o’s diverse, multimodal training enables better generalization to sketch-based tasks, handling semantic ambiguity and visual nuance more effectively.Ultimately, this experiment reveals a key mismatch between the training data of current fashion-specific models and the abstract nature of design sketches, emphasizing the need for more diverse and representative data in fashion model training.

\section{Conclusion}
\label{sec:conclusion_Section}
    We introduce FLORA, the first dataset specifically designed for generating fashion outfits from textual descriptions, alongside NeRA Adapters, a novel architecture leveraging KANs to enhance model adaptability in complex domains. These contributions address two major challenges in AI-driven fashion design: the scarcity of high-quality, semantically rich datasets and the limitations of current techniques in handling fine-grained design control and adaptability. FLORA lays the groundwork for a new generation of generative models capable of interpreting and translating nuanced design prompts into coherent fashion outfit sketches.  Looking ahead, we aim to enhance FLORA with part-level sketch editing, enabling models to modify specific outfit elements while maintaining overall style, offering designers greater control and flexibility. We plan to collaborate with  industry  to broaden the dataset’s diversity. NeRA Adapters, with their nonlinear spline-based design, show strong domain adaptability with modest computational overhead. Future work will focus on optimizing NeRA for efficient, large-scale deployment in real-world fashion applications.

{
    \small
    \bibliographystyle{ieeenat_fullname}
    \bibliography{main}

\begin{thebibliography}{48}
\providecommand{\natexlab}[1]{#1}
\providecommand{\url}[1]{\texttt{#1}}
\expandafter\ifx\csname urlstyle\endcsname\relax
  \providecommand{\doi}[1]{doi: #1}\else
  \providecommand{\doi}{doi: \begingroup \urlstyle{rm}\Url}\fi

\bibitem[fas()]{fashionai2}
\url{https://www.thomasnet.com/insights/zara-h-m-fast-fashion-ai-supply-chain/}.
\newblock [Online; accessed 07-Oct-2024].

\bibitem[Amed et~al.(2022)Amed, Berg, Balchandani, Hedrich, Rölkens, and Young]{mckinsey2022stateoffashion}
Imran Amed, Achim Berg, Vivek Balchandani, Saskia Hedrich, Robb Rölkens, and Kanaiya Young.
\newblock The state of fashion 2022, 2022.
\newblock Accessed: 2024-11-09.

\bibitem[Baldrati et~al.(2023)Baldrati, Morelli, Cartella, Cornia, Bertini, and Cucchiara]{baldrati2023multimodal}
Alberto Baldrati, Davide Morelli, Giuseppe Cartella, Marcella Cornia, Marco Bertini, and Rita Cucchiara.
\newblock Multimodal garment designer: Human-centric latent diffusion models for fashion image editing.
\newblock In \emph{Proceedings of the IEEE/CVF international conference on computer vision}, pages 23393--23402, 2023.

\bibitem[Bodner et~al.(2024)Bodner, Tepsich, Spolski, and Pourteau]{bodner2024convolutional}
Alexander~Dylan Bodner, Antonio~Santiago Tepsich, Jack~Natan Spolski, and Santiago Pourteau.
\newblock Convolutional kolmogorov-arnold networks.
\newblock \emph{arXiv preprint arXiv:2406.13155}, 2024.

\bibitem[Chen et~al.(2024)Chen, Ge, Xie, Wu, Yao, Ren, Wang, Luo, Lu, and Li]{pixart-sigma}
Junsong Chen, Chongjian Ge, Enze Xie, Yue Wu, Lewei Yao, Xiaozhe Ren, Zhongdao Wang, Ping Luo, Huchuan Lu, and Zhenguo Li.
\newblock Pixart-sigma: Weak-to-strong training of diffusion transformer for 4k text-to-image generation, 2024.

\bibitem[Choi et~al.(2021)Choi, Park, Lee, and Choo]{choi2021viton}
Seunghwan Choi, Sunghyun Park, Minsoo Lee, and Jaegul Choo.
\newblock Viton-hd: High-resolution virtual try-on via misalignment-aware normalization.
\newblock In \emph{Proceedings of the IEEE/CVF conference on computer vision and pattern recognition}, pages 14131--14140, 2021.

\bibitem[Esser et~al.(2024)Esser, Kulal, Blattmann, Entezari, Müller, Saini, Levi, Lorenz, Sauer, Boesel, Podell, Dockhorn, English, Lacey, Goodwin, Marek, and Rombach]{Stable-Diffusion-3}
Patrick Esser, Sumith Kulal, Andreas Blattmann, Rahim Entezari, Jonas Müller, Harry Saini, Yam Levi, Dominik Lorenz, Axel Sauer, Frederic Boesel, Dustin Podell, Tim Dockhorn, Zion English, Kyle Lacey, Alex Goodwin, Yannik Marek, and Robin Rombach.
\newblock Scaling rectified flow transformers for high-resolution image synthesis, 2024.

\bibitem[Foong et~al.(2023)Foong, Kotyan, Mao, and Vargas]{foong2023challengesimagegenerationmodels}
Tham~Yik Foong, Shashank Kotyan, Po~Yuan Mao, and Danilo~Vasconcellos Vargas.
\newblock The challenges of image generation models in generating multi-component images, 2023.

\bibitem[Gal et~al.(2022)Gal, Alaluf, Atzmon, Patashnik, Bermano, Chechik, and Cohen-Or]{gal2022image}
Rinon Gal, Yuval Alaluf, Yuval Atzmon, Or Patashnik, Amit~H Bermano, Gal Chechik, and Daniel Cohen-Or.
\newblock An image is worth one word: Personalizing text-to-image generation using textual inversion.
\newblock \emph{arXiv preprint arXiv:2208.01618}, 2022.

\bibitem[Ge et~al.(2019)Ge, Zhang, Wang, Tang, and Luo]{ge2019deepfashion2}
Yuying Ge, Ruimao Zhang, Xiaogang Wang, Xiaoou Tang, and Ping Luo.
\newblock Deepfashion2: A versatile benchmark for detection, pose estimation, segmentation and re-identification of clothing images.
\newblock In \emph{Proceedings of the IEEE/CVF conference on computer vision and pattern recognition}, pages 5337--5345, 2019.

\bibitem[Han et~al.(2018)Han, Wu, Wu, Yu, and Davis]{han2017viton}
Xintong Han, Zuxuan Wu, Zhe Wu, Ruichi Yu, and Larry~S Davis.
\newblock Viton: An image-based virtual try-on network.
\newblock In \emph{CVPR}, 2018.

\bibitem[Han et~al.(2023)Han, Zhu, Yu, Zhang, Song, and Xiang]{han2023fame}
Xiao Han, Xiatian Zhu, Licheng Yu, Li Zhang, Yi-Zhe Song, and Tao Xiang.
\newblock Fame-vil: Multi-tasking vision-language model for heterogeneous fashion tasks.
\newblock In \emph{Proceedings of the IEEE/CVF conference on computer vision and pattern recognition}, pages 2669--2680, 2023.

\bibitem[Hu et~al.(2021)Hu, Shen, Wallis, Allen-Zhu, Li, Wang, Wang, and Chen]{hu2021lora}
Edward~J Hu, Yelong Shen, Phillip Wallis, Zeyuan Allen-Zhu, Yuanzhi Li, Shean Wang, Lu Wang, and Weizhu Chen.
\newblock Lora: Low-rank adaptation of large language models.
\newblock \emph{arXiv preprint arXiv:2106.09685}, 2021.

\bibitem[Hu et~al.(2023)Hu, Wu, Ding, Huang, Deng, and Li]{10213129}
Jianhua Hu, Weimei Wu, Mengjun Ding, Xi Huang, Zhi~Jian Deng, and Xuankai Li.
\newblock A virtual try-on system based on deep learning.
\newblock In \emph{2023 3rd International Symposium on Computer Technology and Information Science (ISCTIS)}, pages 103--107, 2023.

\bibitem[Hyeon-Woo et~al.(2021)Hyeon-Woo, Ye-Bin, and Oh]{hyeon2021fedpara}
Nam Hyeon-Woo, Moon Ye-Bin, and Tae-Hyun Oh.
\newblock Fedpara: Low-rank hadamard product for communication-efficient federated learning.
\newblock \emph{arXiv preprint arXiv:2108.06098}, 2021.

\bibitem[is~Trending This~Year()]{fashionai1}
Fashion AI In 2024:~What is Trending This~Year.
\newblock \url{https://www.intelistyle.com/fashion-ai-in-2023-what-should-we-expect-to-see-this-year/}.
\newblock [Online; accessed 07-Oct-2024].

\bibitem[Isola et~al.(2017)Isola, Zhu, Zhou, and Efros]{pix2pix2017}
Phillip Isola, Jun-Yan Zhu, Tinghui Zhou, and Alexei~A Efros.
\newblock Image-to-image translation with conditional adversarial networks.
\newblock \emph{CVPR}, 2017.

\bibitem[Ivezić and Bagic~Babac(2023)]{surveytext2img}
Dora Ivezić and Marina Bagic~Babac.
\newblock Trends and challenges of text-to-image generation: Sustainability perspective.
\newblock \emph{Croatian Regional Development Journal}, 4:\penalty0 56--77, 2023.

\bibitem[Jiang et~al.(2023)Jiang, Brown, Cheng, Khan, Gupta, Workman, Hanna, Flowers, and Gebru]{10.1145/3600211.3604681}
Harry~H. Jiang, Lauren Brown, Jessica Cheng, Mehtab Khan, Abhishek Gupta, Deja Workman, Alex Hanna, Johnathan Flowers, and Timnit Gebru.
\newblock Ai art and its impact on artists.
\newblock In \emph{Proceedings of the 2023 AAAI/ACM Conference on AI, Ethics, and Society}, page 363–374, New York, NY, USA, 2023. Association for Computing Machinery.

\bibitem[Jiang et~al.(2022)Jiang, Yang, Qiu, Wu, Loy, and Liu]{jiang2022text2human}
Yuming Jiang, Shuai Yang, Haonan Qiu, Wayne Wu, Chen~Change Loy, and Ziwei Liu.
\newblock Text2human: Text-driven controllable human image generation.
\newblock \emph{ACM Transactions on Graphics (TOG)}, 41\penalty0 (4):\penalty0 1--11, 2022.

\bibitem[Kirkpatrick et~al.(2017)Kirkpatrick, Pascanu, Rabinowitz, Veness, Desjardins, Rusu, Milan, Quan, Ramalho, Grabska-Barwinska, et~al.]{kirkpatrick2017overcoming}
James Kirkpatrick, Razvan Pascanu, Neil Rabinowitz, Joel Veness, Guillaume Desjardins, Andrei~A Rusu, Kieran Milan, John Quan, Tiago Ramalho, Agnieszka Grabska-Barwinska, et~al.
\newblock Overcoming catastrophic forgetting in neural networks.
\newblock \emph{Proceedings of the national academy of sciences}, 114\penalty0 (13):\penalty0 3521--3526, 2017.

\bibitem[Labs(2024)]{fluxdev}
Black~Forest Labs.
\newblock {black-forest-labs/FLUX.1-dev · Hugging Face}.
\newblock \url{https://huggingface.co/black-forest-labs/FLUX.1-dev}, 2024.
\newblock [Online; accessed 05-October-2024].

\bibitem[Li et~al.(2024)Li, Liu, Li, Wang, Liu, and Yuan]{li2024u}
Chenxin Li, Xinyu Liu, Wuyang Li, Cheng Wang, Hengyu Liu, and Yixuan Yuan.
\newblock U-kan makes strong backbone for medical image segmentation and generation.
\newblock \emph{arXiv preprint arXiv:2406.02918}, 2024.

\bibitem[Liu et~al.(2023)Liu, Li, Wu, and Lee]{liu2023llava}
Haotian Liu, Chunyuan Li, Qingyang Wu, and Yong~Jae Lee.
\newblock Visual instruction tuning.
\newblock In \emph{NeurIPS}, 2023.

\bibitem[Liu et~al.(2024{\natexlab{a}})Liu, Wang, Yin, Molchanov, Wang, Cheng, and Chen]{liu2024dora}
Shih-Yang Liu, Chien-Yi Wang, Hongxu Yin, Pavlo Molchanov, Yu-Chiang~Frank Wang, Kwang-Ting Cheng, and Min-Hung Chen.
\newblock Dora: Weight-decomposed low-rank adaptation.
\newblock In \emph{Forty-first International Conference on Machine Learning}, 2024{\natexlab{a}}.

\bibitem[Liu et~al.(2016)Liu, Luo, Qiu, Wang, and Tang]{liuLQWTcvpr16DeepFashion}
Ziwei Liu, Ping Luo, Shi Qiu, Xiaogang Wang, and Xiaoou Tang.
\newblock {DeepFashion: Powering Robust Clothes Recognition and Retrieval with Rich Annotations}.
\newblock In \emph{Proceedings of IEEE Conference on Computer Vision and Pattern Recognition (CVPR)}, 2016.

\bibitem[Liu et~al.(2024{\natexlab{b}})Liu, Wang, Vaidya, Ruehle, Halverson, Soljačić, Hou, and Tegmark]{liu2024kankolmogorovarnoldnetworks}
Ziming Liu, Yixuan Wang, Sachin Vaidya, Fabian Ruehle, James Halverson, Marin Soljačić, Thomas~Y. Hou, and Max Tegmark.
\newblock Kan: Kolmogorov-arnold networks, 2024{\natexlab{b}}.

\bibitem[Mirza and Osindero(2014)]{mirza2014conditionalgenerativeadversarialnets}
Mehdi Mirza and Simon Osindero.
\newblock Conditional generative adversarial nets, 2014.

\bibitem[Morales(2019)]{kerasocr}
Fausto Morales.
\newblock {Keras-OCR Documentation}.
\newblock \url{https://keras-ocr.readthedocs.io/en/latest/}, 2019.
\newblock [Online; accessed 05-July-2024].

\bibitem[OpenAI(2024)]{gpt4o}
OpenAI.
\newblock {Hello GPT-4o}.
\newblock \url{https://openai.com/index/hello-gpt-4o/}, 2024.
\newblock [Online; accessed 20-September-2024].

\bibitem[Patel(2024)]{KhushbooPatel2024}
Khushboo Patel.
\newblock Advancements and challenges in text-to-image synthesis: A comprehensive review.
\newblock \emph{International Journal of Intelligent Systems and Applications in Engineering}, 12\penalty0 (3):\penalty0 4228–4237, 2024.

\bibitem[Podell et~al.(2023)Podell, English, Lacey, Blattmann, Dockhorn, Müller, Penna, and Rombach]{SD-XL}
Dustin Podell, Zion English, Kyle Lacey, Andreas Blattmann, Tim Dockhorn, Jonas Müller, Joe Penna, and Robin Rombach.
\newblock Sdxl: Improving latent diffusion models for high-resolution image synthesis, 2023.

\bibitem[Qiu et~al.(2024)Qiu, Zhu, Gong, Chen, and Ning]{qiu2024relu}
Qi Qiu, Tao Zhu, Helin Gong, Liming Chen, and Huansheng Ning.
\newblock Relu-kan: New kolmogorov-arnold networks that only need matrix addition, dot multiplication, and relu.
\newblock \emph{arXiv preprint arXiv:2406.02075}, 2024.

\bibitem[Ramesh et~al.(2021)Ramesh, Pavlov, Goh, Gray, Voss, Radford, Chen, and Sutskever]{ramesh2021zeroshottexttoimagegeneration}
Aditya Ramesh, Mikhail Pavlov, Gabriel Goh, Scott Gray, Chelsea Voss, Alec Radford, Mark Chen, and Ilya Sutskever.
\newblock Zero-shot text-to-image generation, 2021.

\bibitem[Rathore(2019)]{sustfashion}
Bharati Rathore.
\newblock Artificial intelligence in sustainable fashion marketing: Transforming the supply chain landscape.
\newblock \emph{Eduzone : international peer reviewed/refereed academic multidisciplinary journal}, 8:\penalty0 2319--5045, 2019.

\bibitem[Rombach et~al.(2022)Rombach, Blattmann, Lorenz, Esser, and Ommer]{Stable-Diffusion-1.5}
Robin Rombach, Andreas Blattmann, Dominik Lorenz, Patrick Esser, and Bj\"orn Ommer.
\newblock High-resolution image synthesis with latent diffusion models.
\newblock In \emph{Proceedings of the IEEE/CVF Conference on Computer Vision and Pattern Recognition (CVPR)}, pages 10684--10695, 2022.

\bibitem[Rostamzadeh et~al.(2018)Rostamzadeh, Hosseini, Boquet, Stokowiec, Zhang, Jauvin, and Pal]{rostamzadeh2018fashion}
Negar Rostamzadeh, Seyedarian Hosseini, Thomas Boquet, Wojciech Stokowiec, Ying Zhang, Christian Jauvin, and Chris Pal.
\newblock Fashion-gen: The generative fashion dataset and challenge.
\newblock \emph{arXiv preprint arXiv:1806.08317}, 2018.

\bibitem[Ruiz et~al.(2023)Ruiz, Li, Jampani, Pritch, Rubinstein, and Aberman]{ruiz2023dreambooth}
Nataniel Ruiz, Yuanzhen Li, Varun Jampani, Yael Pritch, Michael Rubinstein, and Kfir Aberman.
\newblock Dreambooth: Fine tuning text-to-image diffusion models for subject-driven generation.
\newblock In \emph{Proceedings of the IEEE/CVF conference on computer vision and pattern recognition}, pages 22500--22510, 2023.

\bibitem[Saharia et~al.(2022)Saharia, Chan, Saxena, Li, Whang, Denton, Ghasemipour, Ayan, Mahdavi, Lopes, Salimans, Ho, Fleet, and Norouzi]{saharia2022photorealistictexttoimagediffusionmodels}
Chitwan Saharia, William Chan, Saurabh Saxena, Lala Li, Jay Whang, Emily Denton, Seyed Kamyar~Seyed Ghasemipour, Burcu~Karagol Ayan, S.~Sara Mahdavi, Rapha~Gontijo Lopes, Tim Salimans, Jonathan Ho, David~J Fleet, and Mohammad Norouzi.
\newblock Photorealistic text-to-image diffusion models with deep language understanding, 2022.

\bibitem[Saponaro et~al.(2018)Saponaro, Le~Gal, Gao, Guisiano, and Maniere]{8601258}
Mariapaola Saponaro, Diane Le~Gal, Manjiao Gao, Matthieu Guisiano, and Ivan~Coste Maniere.
\newblock Challenges and opportunities of artificial intelligence in the fashion world.
\newblock In \emph{2018 International Conference on Intelligent and Innovative Computing Applications (ICONIC)}, pages 1--5, 2018.

\bibitem[schroneko(2024)]{transparent-background}
schroneko.
\newblock {Transparent Background - a Hugging Face Space by schroneko}.
\newblock \url{https://huggingface.co/spaces/schroneko/transparent-background}, 2024.
\newblock [Online; accessed 05-July-2024].

\bibitem[Schuhmann et~al.(2022)]{schuhmann2022laion}
Christoph Schuhmann et~al.
\newblock Laion-5b: An open large-scale dataset for training next generation image-text models.
\newblock \emph{NeurIPS}, 2022.

\bibitem[Susladkar(2024)]{scenetexteraser}
Onkar Susladkar.
\newblock {Controlnet Stable diffusion scene text Eraser}.
\newblock \url{https://huggingface.co/onkarsus13/controlnet_stablediffusion_scenetextEraser}, 2024.
\newblock [Online; accessed 25-September-2024].

\bibitem[Yang and Wang(2024)]{yang2024kolmogorov}
Xingyi Yang and Xinchao Wang.
\newblock Kolmogorov-arnold transformer.
\newblock \emph{arXiv preprint arXiv:2409.10594}, 2024.

\bibitem[Yeh et~al.(2023)Yeh, Hsieh, Gao, Yang, Oh, and Gong]{yeh2023navigating}
Shih-Ying Yeh, Yu-Guan Hsieh, Zhidong Gao, Bernard~BW Yang, Giyeong Oh, and Yanmin Gong.
\newblock Navigating text-to-image customization: From lycoris fine-tuning to model evaluation.
\newblock In \emph{The Twelfth International Conference on Learning Representations}, 2023.

\bibitem[Yirui et~al.(2018)Yirui, Liu, Gao, and Su]{fashionGAN}
Cui Yirui, Q. Liu, C. Gao, and Zhuo Su.
\newblock Fashiongan: Display your fashion design using conditional generative adversarial nets.
\newblock \emph{Computer Graphics Forum}, 37:\penalty0 109--119, 2018.

\bibitem[Zhao et~al.(2024)Zhao, Zhang, Zhang, and Wu]{zhao2024unifashion}
Xiangyu Zhao, Yuehan Zhang, Wenlong Zhang, and Xiao-Ming Wu.
\newblock Unifashion: A unified vision-language model for multimodal fashion retrieval and generation.
\newblock \emph{arXiv preprint arXiv:2408.11305}, 2024.

\bibitem[Zheng et~al.(2015)Zheng, Shen, Tian, Wang, Wang, and Tian]{Zheng_2015_ICCV}
Liang Zheng, Liyue Shen, Lu Tian, Shengjin Wang, Jingdong Wang, and Qi Tian.
\newblock Scalable person re-identification: A benchmark.
\newblock In \emph{Proceedings of the IEEE International Conference on Computer Vision (ICCV)}, 2015.

\end{thebibliography}
}

\clearpage
\setcounter{page}{1}
\maketitlesupplementary

\begin{figure*}[ht]
          \centering
          \includegraphics[width=0.9\textwidth]{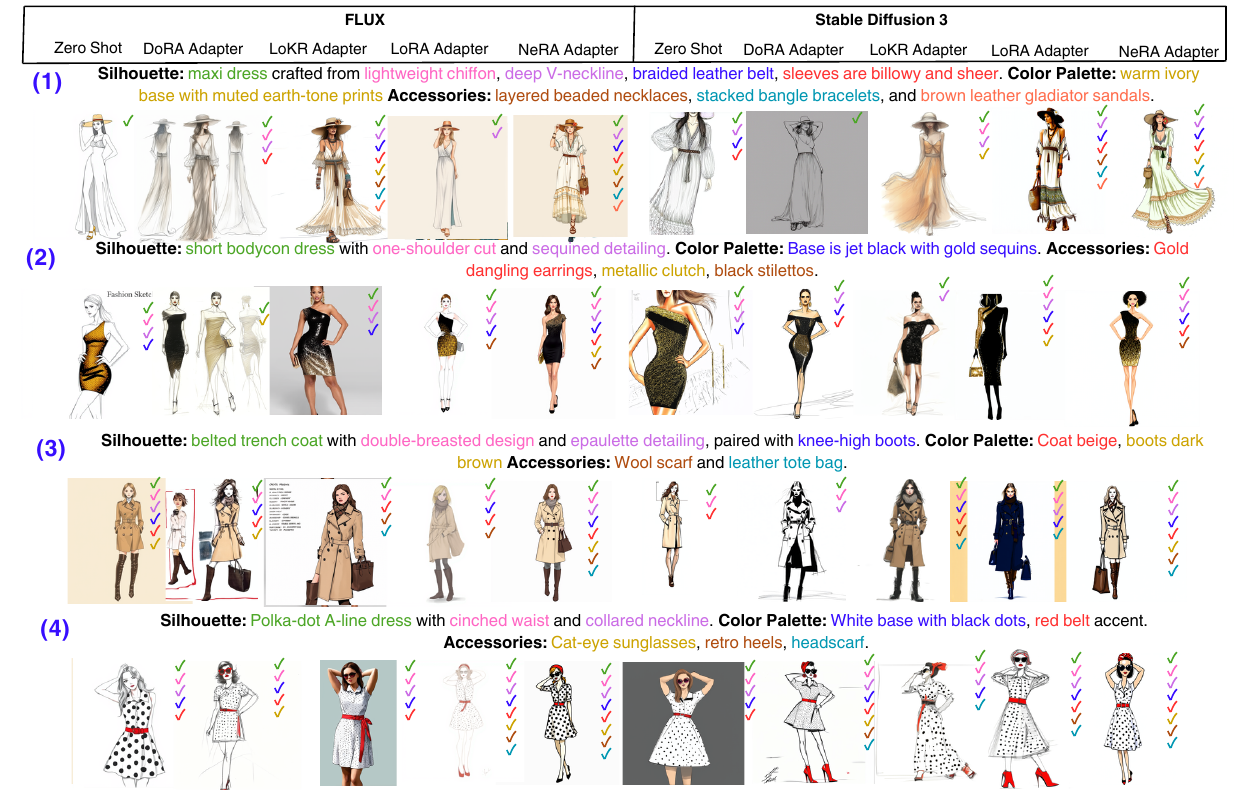}
          \caption{ Additional qualitative results comparing NeRA, LoRA, and zero-shot performance. Colored arrows mark whether the prompt elements written in that color were correctly generated.}  
          \label{fig:additinal_results}
\end{figure*}
This supplementary material includes the following sections:

\begin{itemize}
 \item  \textbf{Additional Qualitative Results}  
\item \textbf{Human Evaluation Results}  
 \item \textbf{Ablation Study} 
\item \textbf{FLORA Dataset} 
\item \textbf{Fine-Tuning Template for Outfit Sketch Generation} 
\item \textbf{Implementation and FLORA Test Set Details}

\item \textbf{NeRA vs. LoRA: A Mathematical Justification}
\end{itemize}

\section{Additional Qualitative Results}
Figure \ref{fig:additinal_results} provides additional qualitative comparison of zero-shot, MLP-based adapters and the proposed NeRA using FLUX and Stable Diffusion 3 as baselines. These examples illustrate model performance on a variety of fashion prompts, covering diverse silhouettes, color schemes, and accessory details.

\section{Human Evaluation Results}
    \textbf{Human Evaluation Metrics}
    For a thorough evaluation of the generated fashion sketches, we defined seven key metrics, each capturing specific aspects of image quality and alignment with the prompt. Table \ref{tab:human_eval} presents human evaluation results. These metrics allowed us to perform a comprehensive assessment, ensuring that both technical quality and alignment with the prompt are rigorously evaluated: 
    \begin{itemize}
        \item \textbf{Prompt Relevance:}  
        Evaluates how accurately the generated image reflects the essential elements described in the prompt. This metric focuses on effectively visualizing the main idea or theme specified.
    
        \item \textbf{Visual Quality:}  
        Assesses the technical quality of the image, including factors such as resolution, sharpness, clarity, and the overall visual appeal.
    
        \item \textbf{Creativity:}  
        Measures the degree of innovation and imaginative interpretation in the image. This metric considers how the generated image demonstrates creativity while staying coherent with the prompt's details.
    
        \item \textbf{Pose and Model Accuracy:}  
        Examines the alignment between the model's physical representation in the image, such as pose, gender, and stance and the specifications given in the prompt.
    
        \item \textbf{Color Palette Alignment:}  
        Compares the colors used in the generated image to the colors described in the prompt, assessing accuracy in terms of color scheme and consistency.
    
        \item \textbf{Contextual Fit:}  
        Evaluates how well the image fits the broader context implied by the prompt, including aspects like mood, setting, and style.
    
        \item \textbf{Overall Appeal:}  
        Provides a holistic assessment of the image's attractiveness and visual impact, taking into account its general appeal to viewers.
    \end{itemize}

\section{Ablation Study}

    \begin{table}[]
        \centering
        \setlength{\tabcolsep}{10pt}
        \begin{tabular}{ccc}
        \hline         
        \multicolumn{3}{c}{\textit{LoRA Rank}}                                                                                         \\ \hline
        \textbf{LoRA Rank}                                               & \textbf{FID $\downarrow$}   & \textbf{CLIP SIM $\uparrow$}                          \\ \hline
        $r = 8$                                                          & 8.76                        & 0.2711                                                \\
        $r = 16$                                                         & 8.13                        & 0.2832                                                \\
        $r = 32$                                                         & 7.92                        & 0.2901                                                \\
        $r = 64$ (Default)                                               & \textbf{7.88}               & \textbf{0.2987}                                       \\ \hline        
        \multicolumn{3}{c}{\textit{NeRA's Downsample Rate}}                                                                               \\ \hline
        \textbf{Downsample Rate}                                         & \textbf{FID $\downarrow$}   & \textbf{CLIP SIM $\uparrow$}                          \\ \hline
        $m/16$                                                           & 7.18                        & 0.3298                                                \\
        $m/8$                                                            & 6.66                        & 0.3319                                                \\
        $m/4$                                                            & 6.49                        & 0.3401                                                \\
        $m/2$ (Default)                                                  & \textbf{6.05}               & \textbf{0.3412}                                       \\ \hline        
        \multicolumn{3}{c}{\textit{Finetuning Techniques}}                                                                             \\ \hline
                                                                         & \textbf{FID $\downarrow$}   & \textbf{CLIP SIM $\uparrow$}                          \\ \hline        
        {\color[HTML]{222222} DreamBooth}                                & {\color[HTML]{000000} 8.37} & {\color[HTML]{000000} 0.3217}                         \\
        \color[HTML]{222222} Textual Inversion & {\color[HTML]{000000} 8.88} & \color[HTML]{222222} 0.3158 \\ \hline
        \end{tabular}
        \caption{Ablation study comparing the effect of varying ranks and downsample rate for LoRA and NeRA respectively, also different finetuning techniques using FLUX as the baseline.}
        \label{tab:ablation_lora_kan}
    \end{table}

    In this section, we investigate the use of NeRA adapters as an alternative to LoRA. Extending our previous ablation studies, we explore the effect of varying ranks for LoRA and downsample rates for NeRA. For the ablation, we choose FLUX as our baseline model for conducting the experiments. Table \ref{tab:ablation_lora_kan} provides the CLIP-SIM and FID scores for different ranks and downsampling rates for the respective adapters.

    \textbf{Effect of Rank on LoRA Performance:} For LoRA, we observe a progressive improvement in both CLIP-SIM and FID scores as the rank increases. The model achieves the highest performance at rank $r=64$, yielding a CLIP-SIM of 0.2987 and an FID of 7.88. This indicates that increasing the rank allows LoRA to better capture domain-specific details relevant to our sketch generation task.

    \textbf{Effect of Downsample Rate on NeRA's Performance:} The downsample rate in the NeRA refers to the dimensionality reduction applied to the input features. Specifically, as explained in proposed methodology section, the NeRA incorporates two main functions: the $\phi_{down}$ and $\phi_{up}$. $\phi_{down}$ projects the input features from the original dimensionality ($m$) to a reduced dimensionality($n$), and the $\phi_{up}$ projects the reduced-dimensional features back to their original size $m$. Here, $n$ is the downsampling rate, which is hyper-parameter and is decided by the ratio by which we want to reduce dimension (e.g. $m/16$, $m/8$, $m/4$, $m/2$). Where a smaller downsample rate ($m/16$) implies greater reduction and larger rates ($m/2$) retain more of the original feature space.
    
    The ablation study, as shown in the table \ref{tab:ablation_lora_kan}, evaluates the performance of NeRA at different downsample rates. As the downsampling rate increases from $m/16$ to $m/2$ results in improvements in both CLIP-SIM and FID scores. This progression highlights the trade-off between dimensionality reduction and feature retention, with higher downsample rates striking an optimal balance between computational efficiency and model performance. These findings emphasize the importance of selecting an appropriate downsample rate to maximize the NeRA's effectiveness.

    \textbf{Comparison with Existing Personalization Methods:} To assess the strengths of our adapter-based fine-tuning approach, we compare it against two widely used personalization techniques - DreamBooth \cite{ruiz2023dreambooth} and Textual Inversion \cite{gal2022image} each representing different ends of the efficiency fidelity trade-off. Both methods were evaluated using the Flux variant of our model.
    DreamBooth achieved a FID of 8.37 and a CLIP-SIM of 0.3217, while Textual Inversion scored 8.88 and 0.3158, respectively. Although both methods outperform zero-shot baselines, they fall short of the performance achieved by Flux fine-tuned with our LoRA and NeRA (see Table 1 from main paper).
    DreamBooth, despite its high-fidelity generations, tends to overfit in few-shot settings and struggles with abstract or compositional prompts. Textual Inversion, on the other hand, lacks the flexibility needed for task-specific or multi-domain generation. In contrast, adapter-based approach, featuring multi-stage adaptation and prompt-aligned tuning more effectively captures compositional semantics and delivers superior generalization across tasks.

\section{FLORA Dataset} 
\subsection{Ethical and legal considerations in use of FLORA} 
The images included in FLORA were sourced from publicly available platforms. This dataset is strictly intended for non-commercial, research purposes, and its use must comply with copyright and intellectual property laws.

    \subsection{Dataset Creation and Filtering Process}
        To create the FLORA dataset, we employed a multi-stage filtering and description process to ensure that the images met quality standards and contained only the desired fashion elements. Below, we describe the stages of filtering, noise removal, and detailed description generation, along with the prompts used at each step.

\textbf{Initial Filtering of Noise Using VLM Model:} In the initial filtering stage, we used a VLM model (LLaVa 32b) to identify and remove images that did not meet the criteria for inclusion in the dataset. Specifically, we aimed to retain images that contained a single outfit sketch without extraneous objects. To achieve this, we provided each image with the following prompts and used the model's responses to segregate the data:
        \begin{promptbox}
            \textbf{Prompt 1:} \textit{``Is there a single outfit present in the image? Yes or No?" }\\
            \textbf{Prompt 2:} \textit{``Is the outfit present in the image a sketch? Yes or No?" }\\
            \textbf{Prompt 3:} \textit{``Does the image only contain the sketch without any other objects? Yes or No?" }\\ 
            \textbf{Prompt 4:} \textit{``Are there multiple outfits present in the image? Yes or No?"}
        \end{promptbox}
        If the VLLM model answered ``Yes" to the questions indicating the desired characteristics (single outfit, sketch format, no additional objects, and no multiple outfits), the image proceeded to the next filtering stage. If the answer to any of these prompts was "No," the image was discarded.\\

        \textbf{Watermark Detection and Removal:} For images that passed the initial filtering, we implemented an additional check to detect and remove watermarks, which could interfere with model training. At this stage as well we utilized the LLaVa 32b with the following prompt:
        \begin{promptbox}
            \textbf{Prompt:} \textit{``Does this image contain a watermark? A watermark may appear as text or a logo superimposed over the image. Yes or No"}
        \end{promptbox}
        If the model identified a watermark ("Yes" response), we performed image inpainting to remove the watermark, as described in the main paper. This ensured that the images in the dataset were clean and free from visual obstructions.\\

        \textbf{Input Prompt Generation:} Once the images were filtered, we used GPT-4o to generate precise and comprehensive descriptions of each outfit. This process involved a multi-turn approach, where each turn focused on describing a specific aspect of the image. The following prompts were used to extract detailed information:
            \begin{promptbox}
            \textbf{Model Pose Description:} ``Describe the model's posture. Is the model standing, sitting, walking, or in any other specific pose? Detail the position of the model's hands, legs, and overall body stance. Provide enough detail so that someone reading the description could recreate the sketch with high accuracy. Note: Generate a response in a single paragraph with not more than 50 words. Please note that - while describing pose, do not describe the dress the model is wearing or do not use any adjectives. Just in simple words, explain the pose of the model, nothing else."
            \end{promptbox}

            \begin{promptbox}
            \textbf{Outfit Description:} ``Describe the outfit present in the image. Describe in a way that a fashion designer can read that description and create an exact similar sketch of the outfit. Use fashion design terminology. Describe garment type, fabric, cut, and any special features. Note: Generate a response in a single paragraph with not more than 200 words."
            \end{promptbox}

            \begin{promptbox}
            \textbf{Color Details:} ``Provide a detailed description of the color scheme used in the outfit. Mention the primary color, any secondary colors, and any unique color patterns. Note: Generate the response in a single paragraph with not more than 50 words. Just return color-related information."
            \end{promptbox}
            
            \begin{promptbox}
            \textbf{Accessories Description:} ``Mention and describe any accessories or additional design elements included in the sketch, such as belts, shoes, jewelry, or headpieces. Provide enough detail so that someone reading the description could recreate the sketch with high accuracy. Note: Generate the response in a single paragraph with not more than 50 words. If no accessories are present, respond with `no additional accessories present'."
            \end{promptbox}

This multi-turn approach allowed us to capture specific details of each image, resulting in a dataset that accurately reflects the visual and stylistic characteristics required for high-quality, fashion-focused AI training.

Additionally, we extracted information on the presence and gender of the model in each image using LLaVa 32b. By using this layered approach - initial filtering, watermark detection, and detailed multi-turn descriptions; we created a high-quality, fashion-specific dataset (FLORA) that supports the generation of accurate and stylistically rich images from textual descriptions. This structured dataset creation process ensures that the FLORA dataset is diverse, precise, and highly applicable to fashion-focused generative AI tasks.

    \subsection{Dataset Insights and Diversity Analysis}
\begin{figure}[ht]
          \centering
          \includegraphics[width=0.5\textwidth]{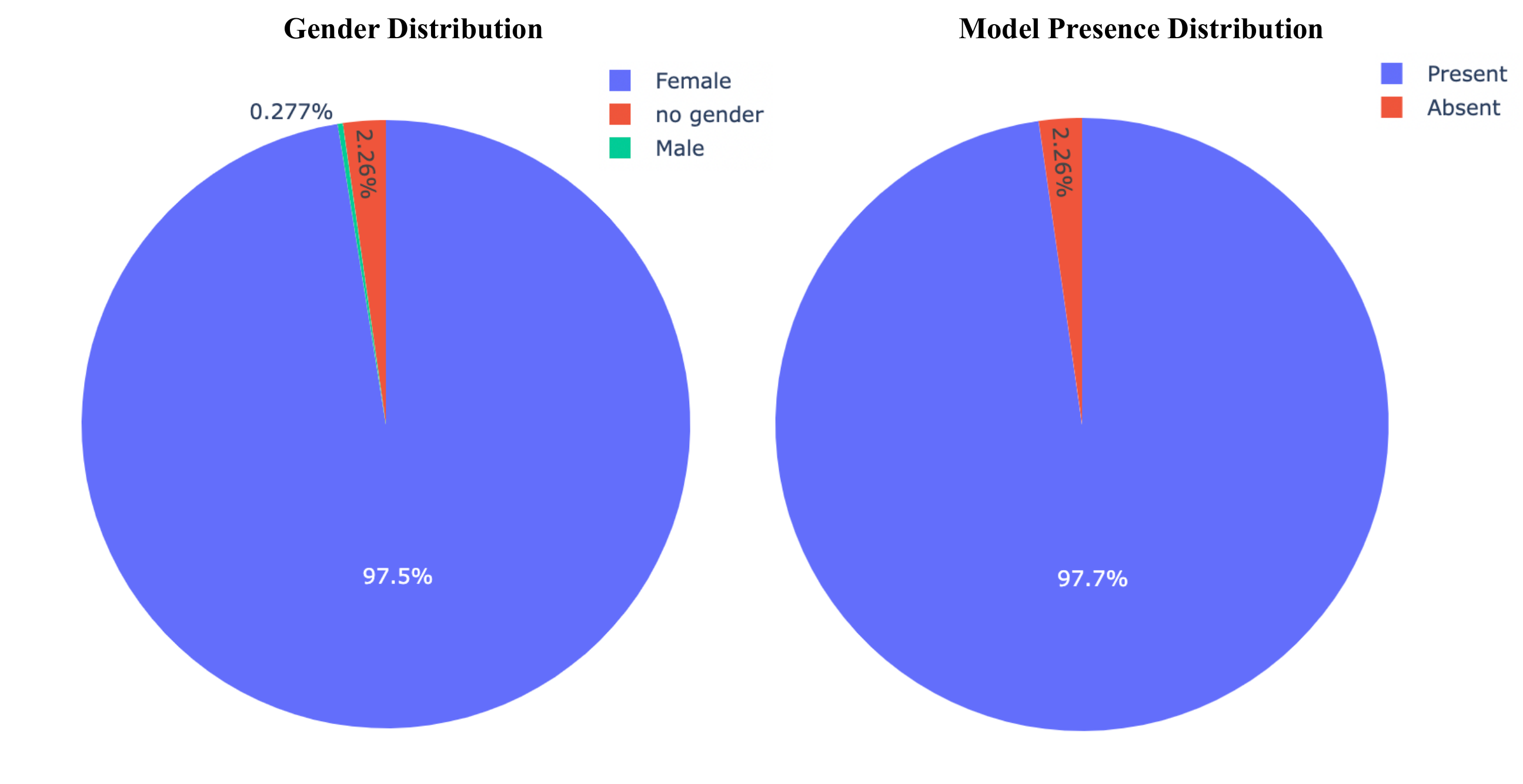}
          \caption{Distribution of gender representation and model presence in the FLORA dataset.} 
          \label{fig:model_gender_pie_chart}
        \end{figure}

        Figure \ref{fig:model_gender_pie_chart} summarizes the gender representation and model presence in the FLORA dataset. Of the images, 97.5\% (4,220) feature female outfits, 0.27\% (12) male outfits, and 2.26\% (98) show "no gender," where the outfit is displayed without a model. Similarly, 97\% of the images include a model wearing the outfit, while 3\% display only the outfit. These distributions in the FLORA dataset are influenced by the types of fashion images commonly found on the internet. For example, the high percentage of female outfits and model presence results from the dataset being sourced from online data, where such trends are prevalent.\\      

\begin{figure*}
            \centering
            \includegraphics[width=1\linewidth]{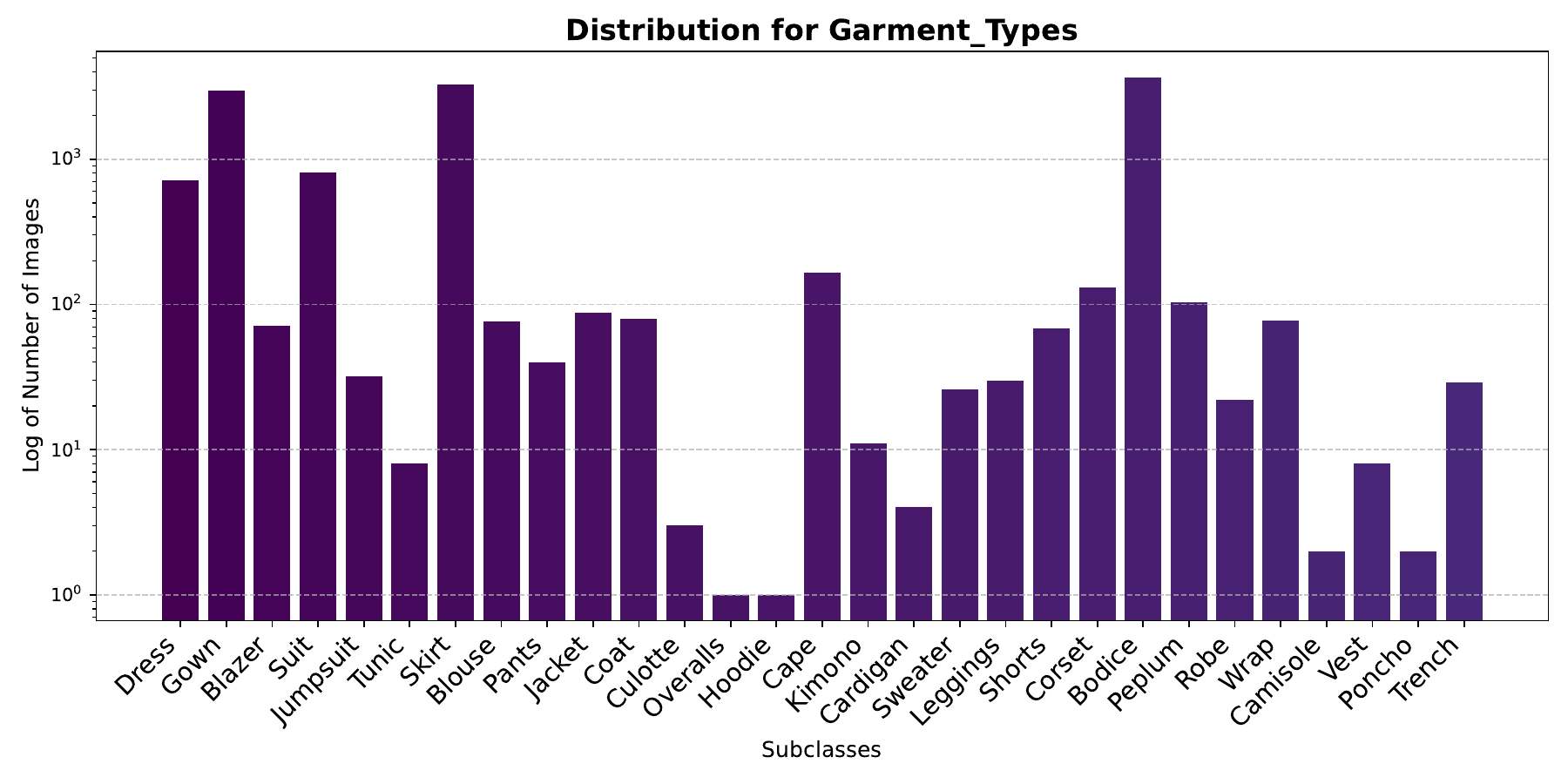}
            \caption{Image Distribution for 29 Garment Types available in FLORA}
            \label{fig:lg1}
        \end{figure*}

\begin{figure*}
            \centering
            \includegraphics[width=1\linewidth]{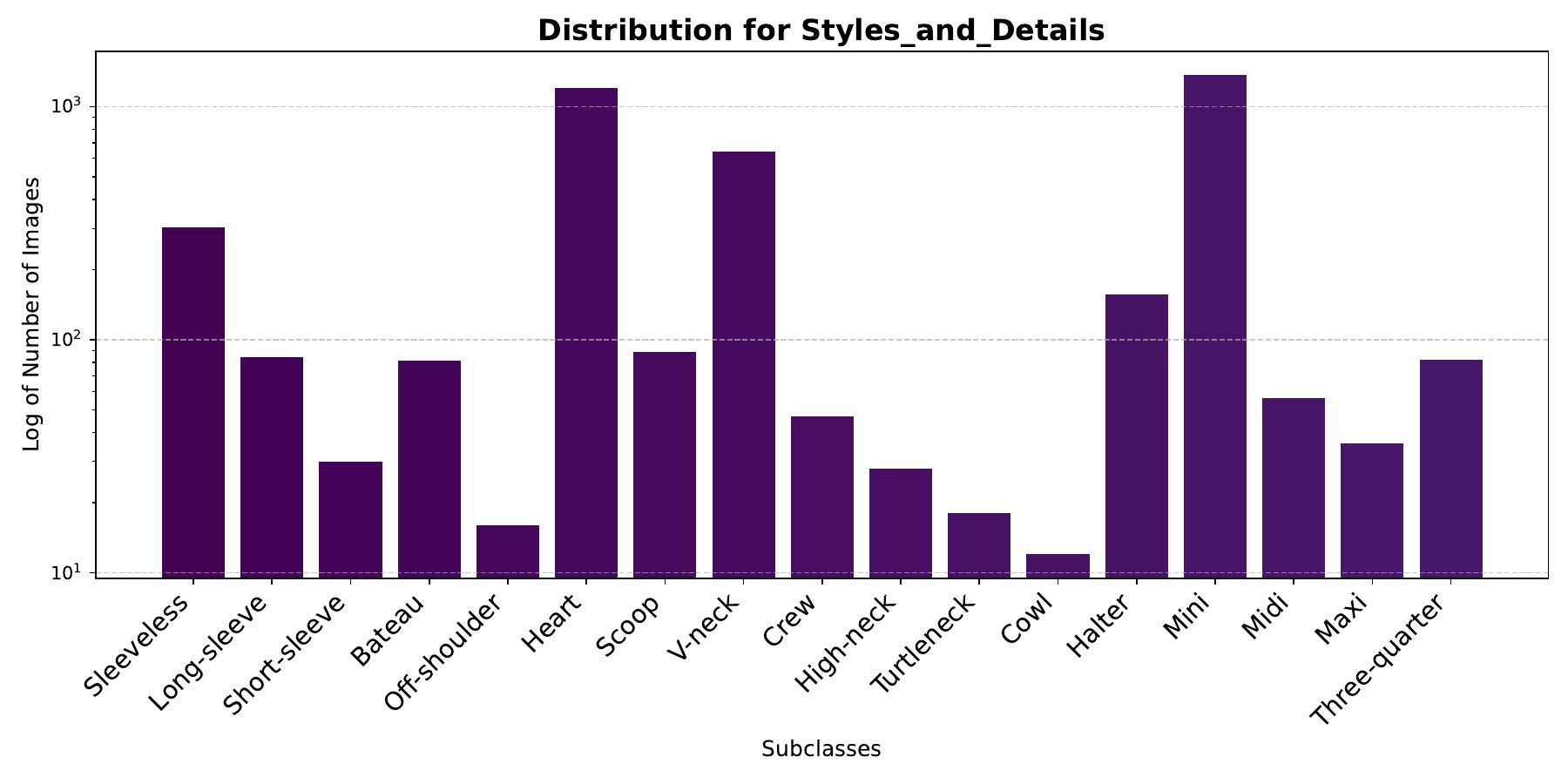}
            \caption{Image Distribution for 17 Styles}
            \label{fig:lg2}
        \end{figure*}

\begin{figure*}
            \centering
            \includegraphics[width=1\linewidth]{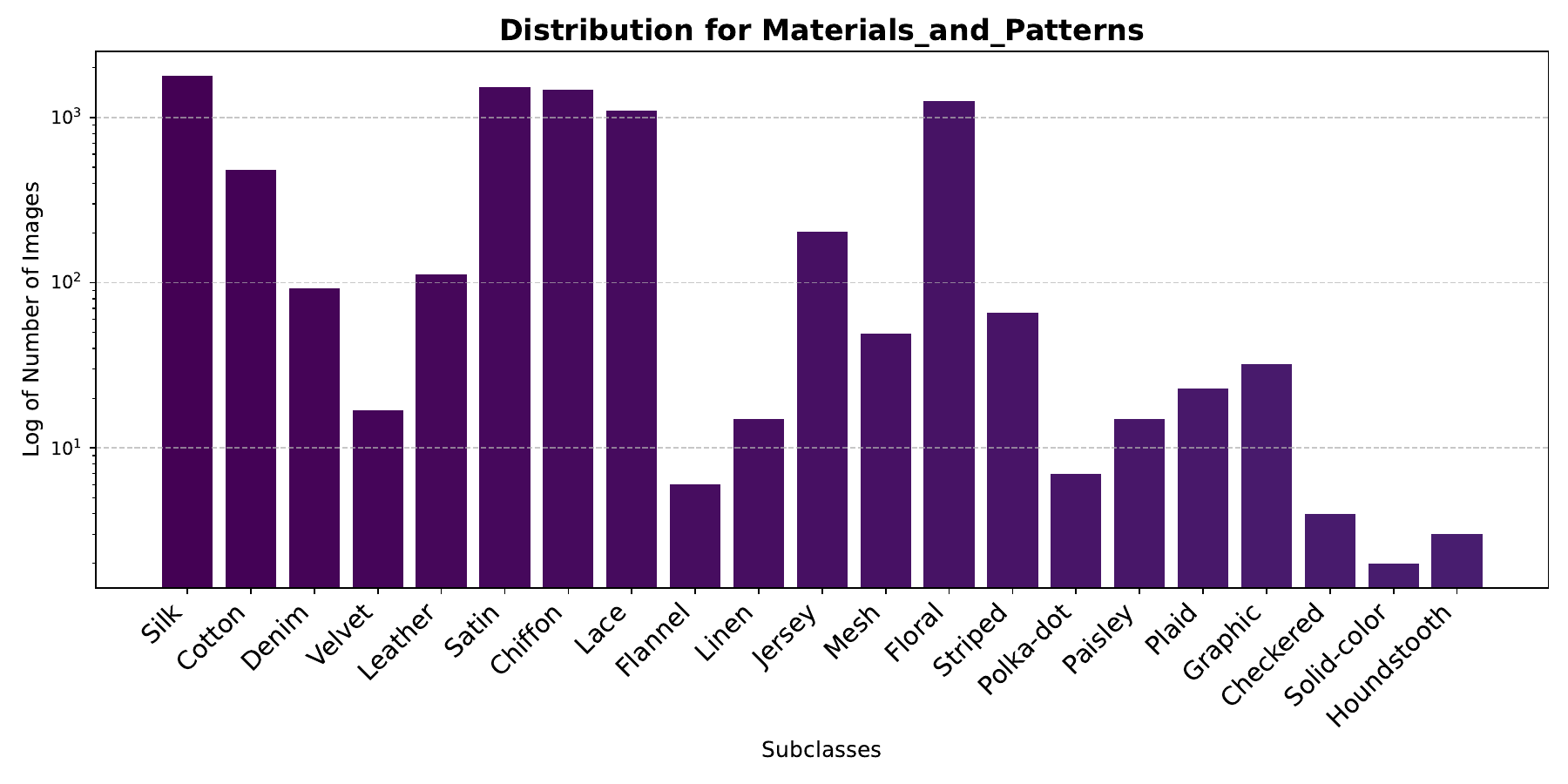}
            \caption{Image Distribution for 21 Materials \& Patterns}
            \label{fig:lg3}
        \end{figure*}

\begin{figure*}
            \centering
            \includegraphics[width=1\linewidth]{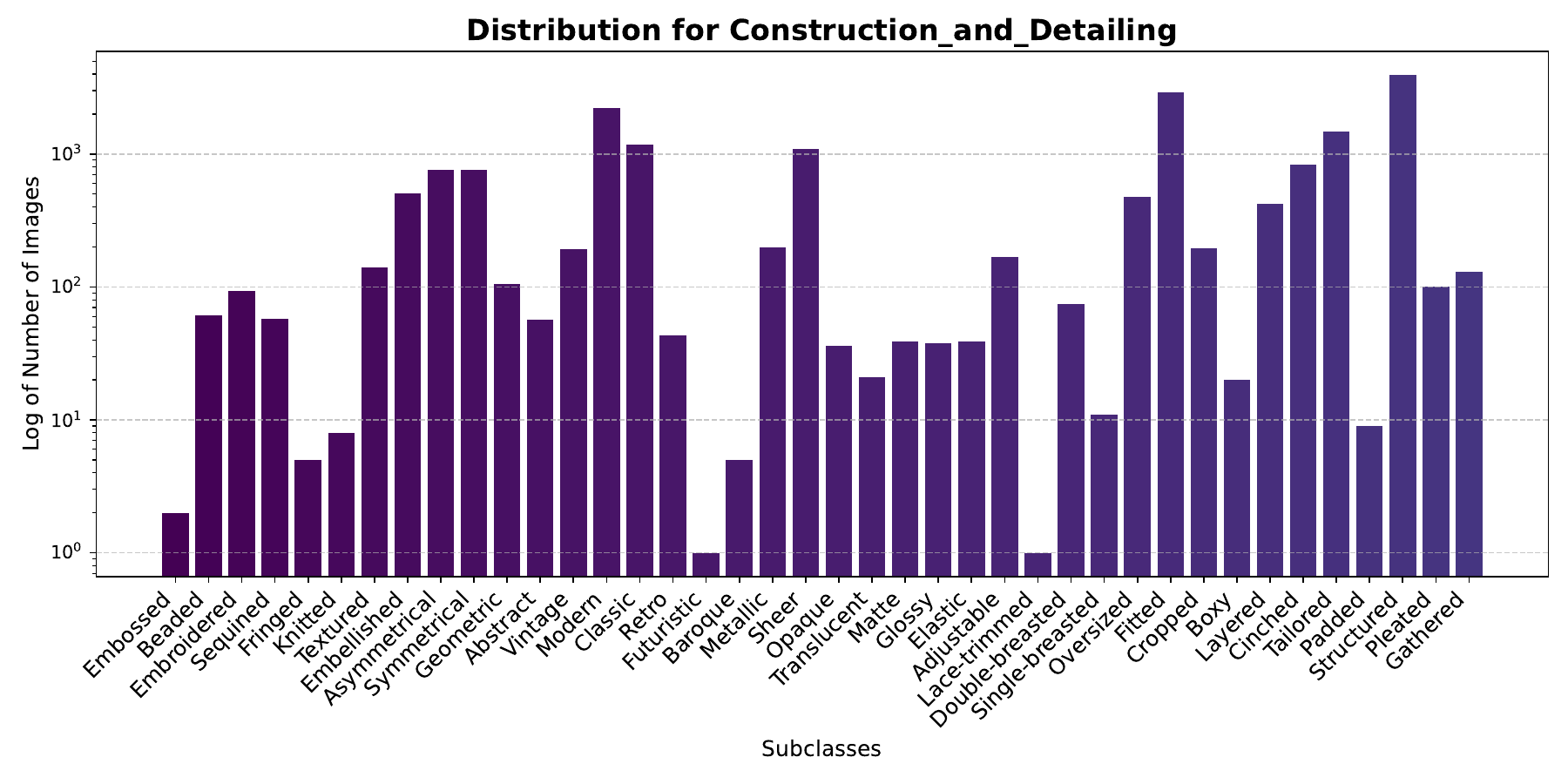}
            \caption{Image Distribution for 40 Construction styles}
            \label{fig:lg4}
        \end{figure*}

\begin{figure*}
            \centering
            \includegraphics[width=1\linewidth]{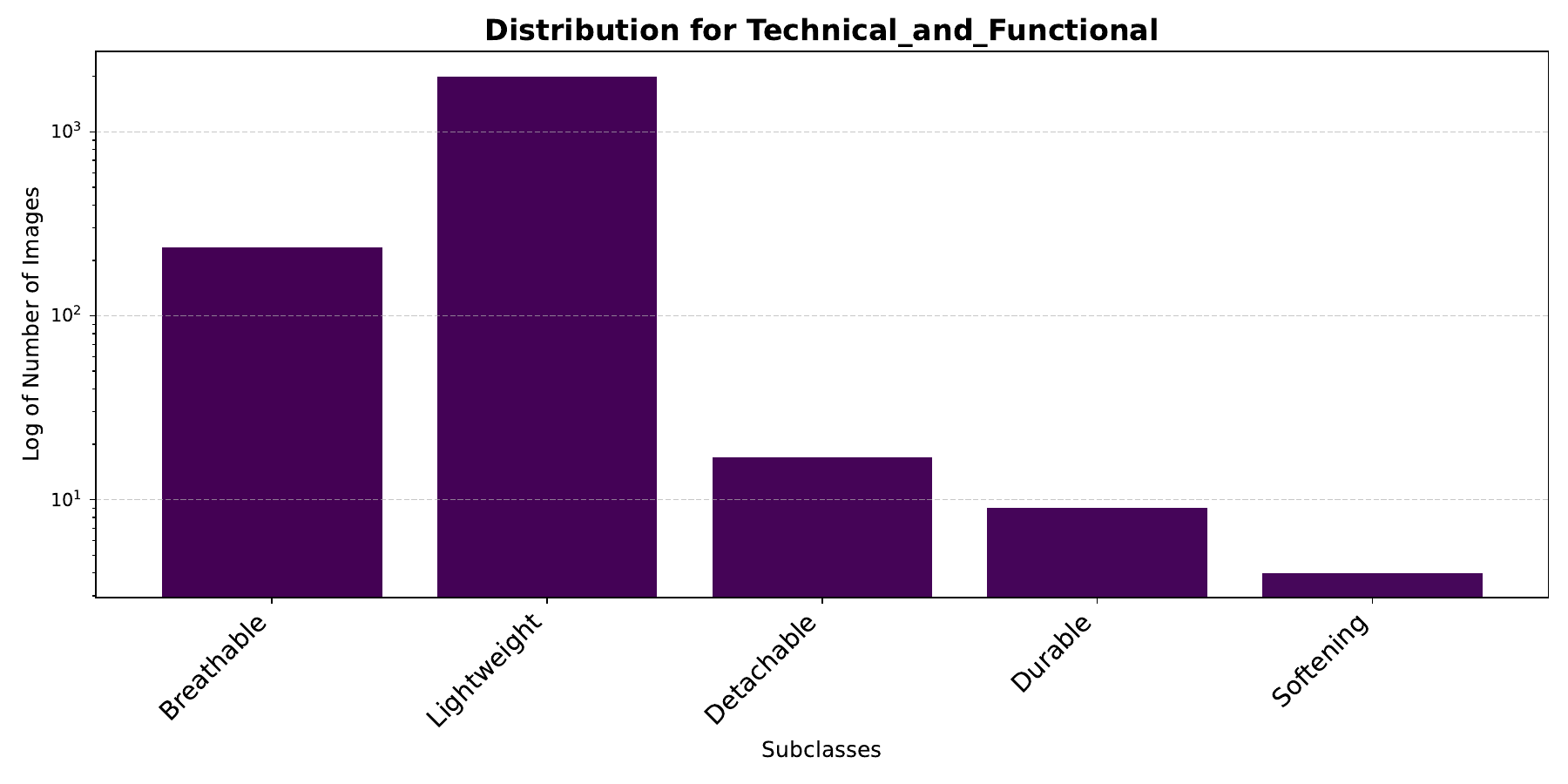}
            \caption{Image Distribution for Technical styles}
            \label{fig:lg5}
        \end{figure*}

\begin{figure*}
            \centering
            \includegraphics[width=1\linewidth]{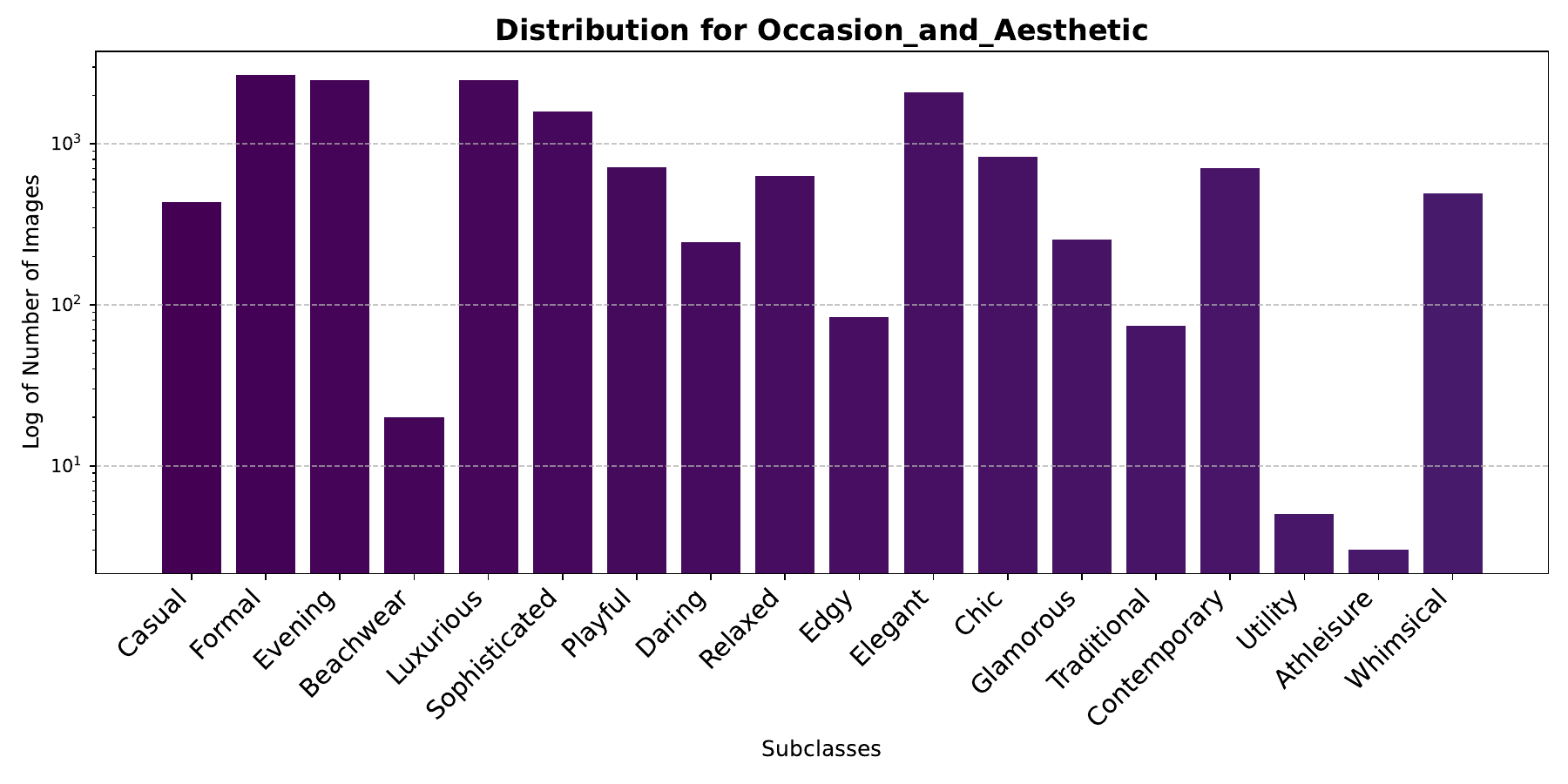}
            \caption{Image Distribution for 18 Occasion \& Aesthetic styles}
            \label{fig:lg6}
        \end{figure*}

\begin{figure*}
            \centering
            \includegraphics[width=1\linewidth]{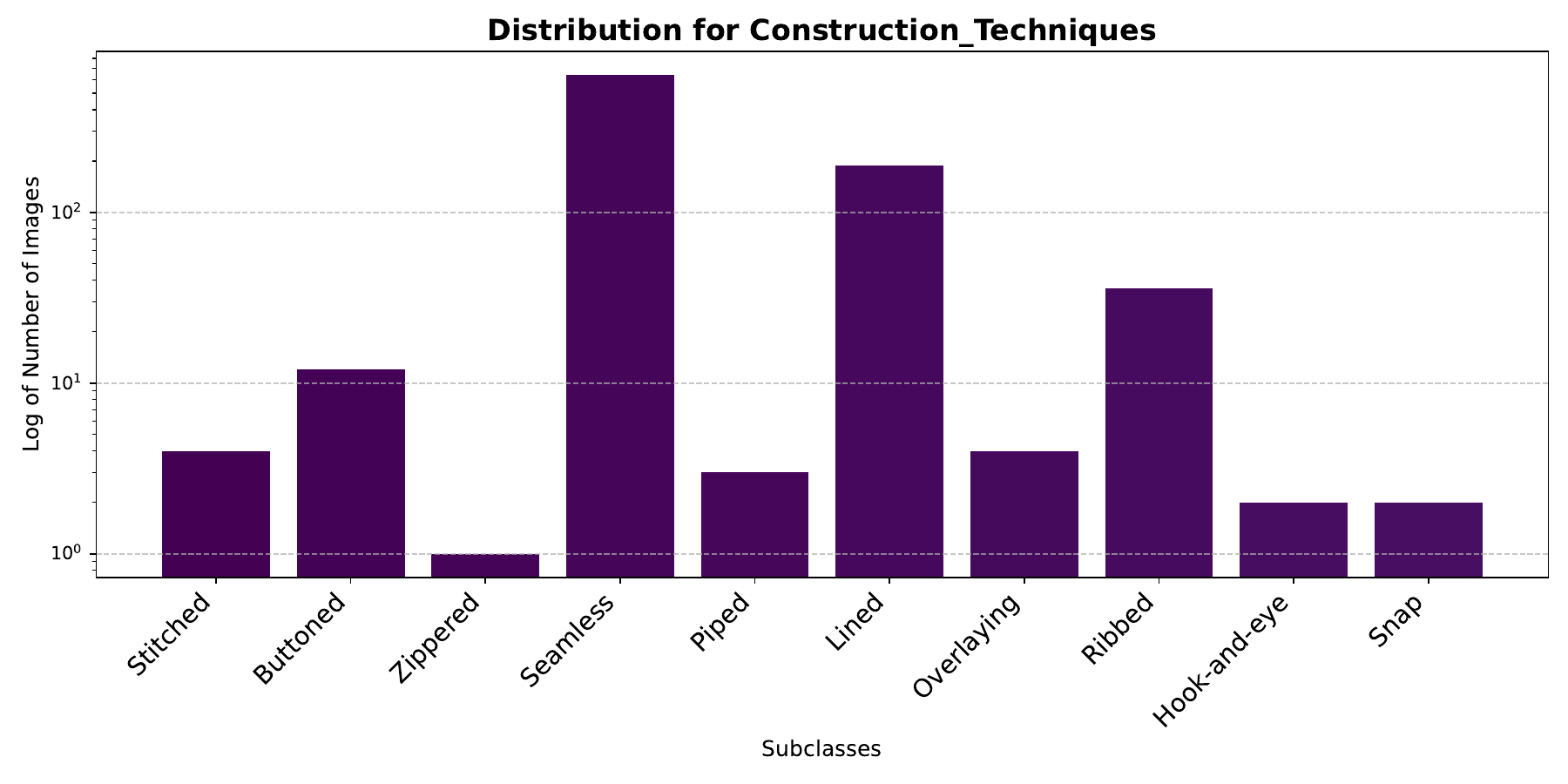}
            \caption{Image Distribution for 10 Construction techniques}
            \label{fig:lg7}
        \end{figure*}

\begin{figure*}
            \centering
            \includegraphics[width=1\linewidth]{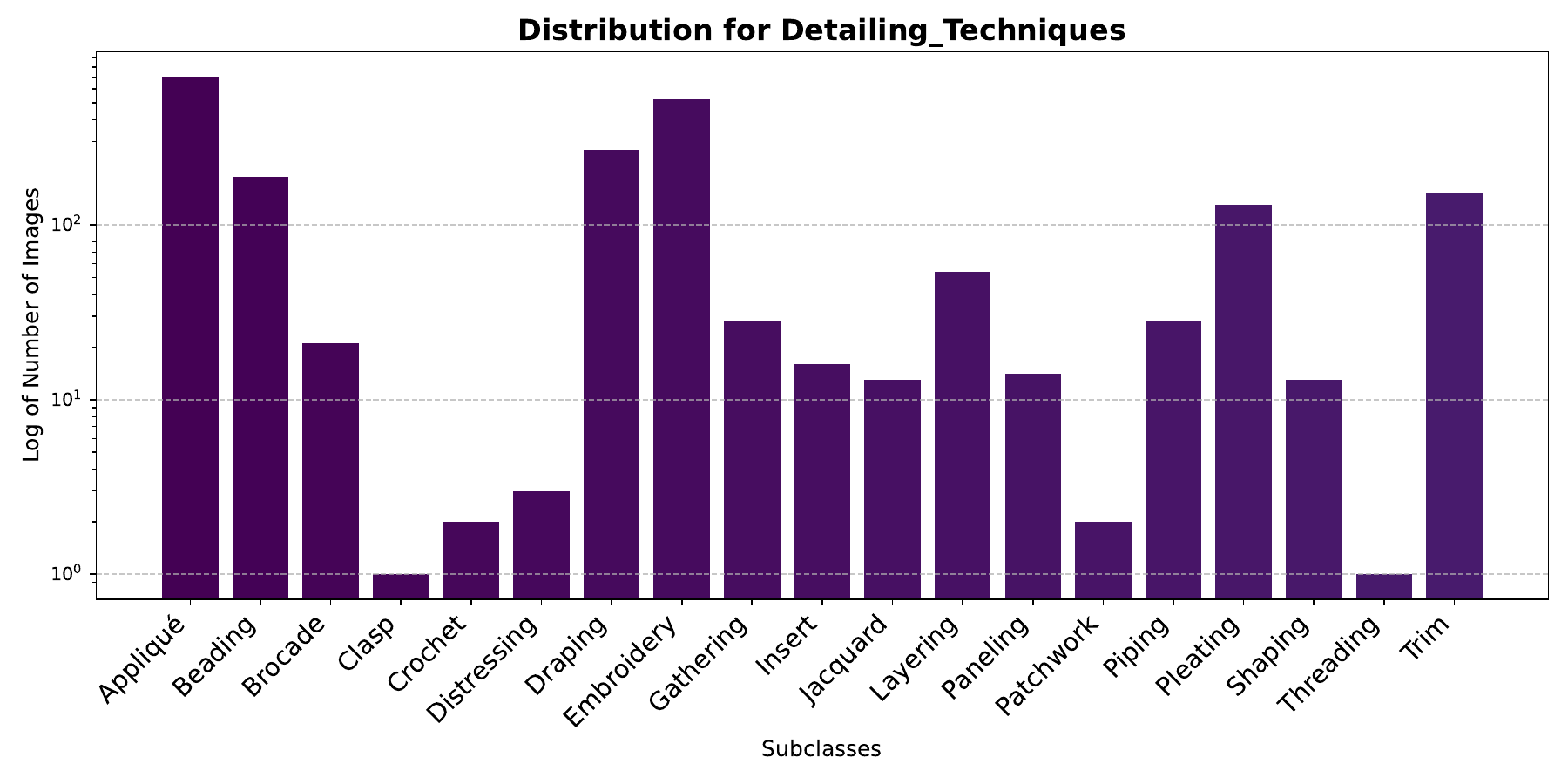}
            \caption{Image Distribution for 19 Detailing techniques}
            \label{fig:lg8}
        \end{figure*}

\begin{figure*}
            \centering
            \includegraphics[width=1\linewidth]{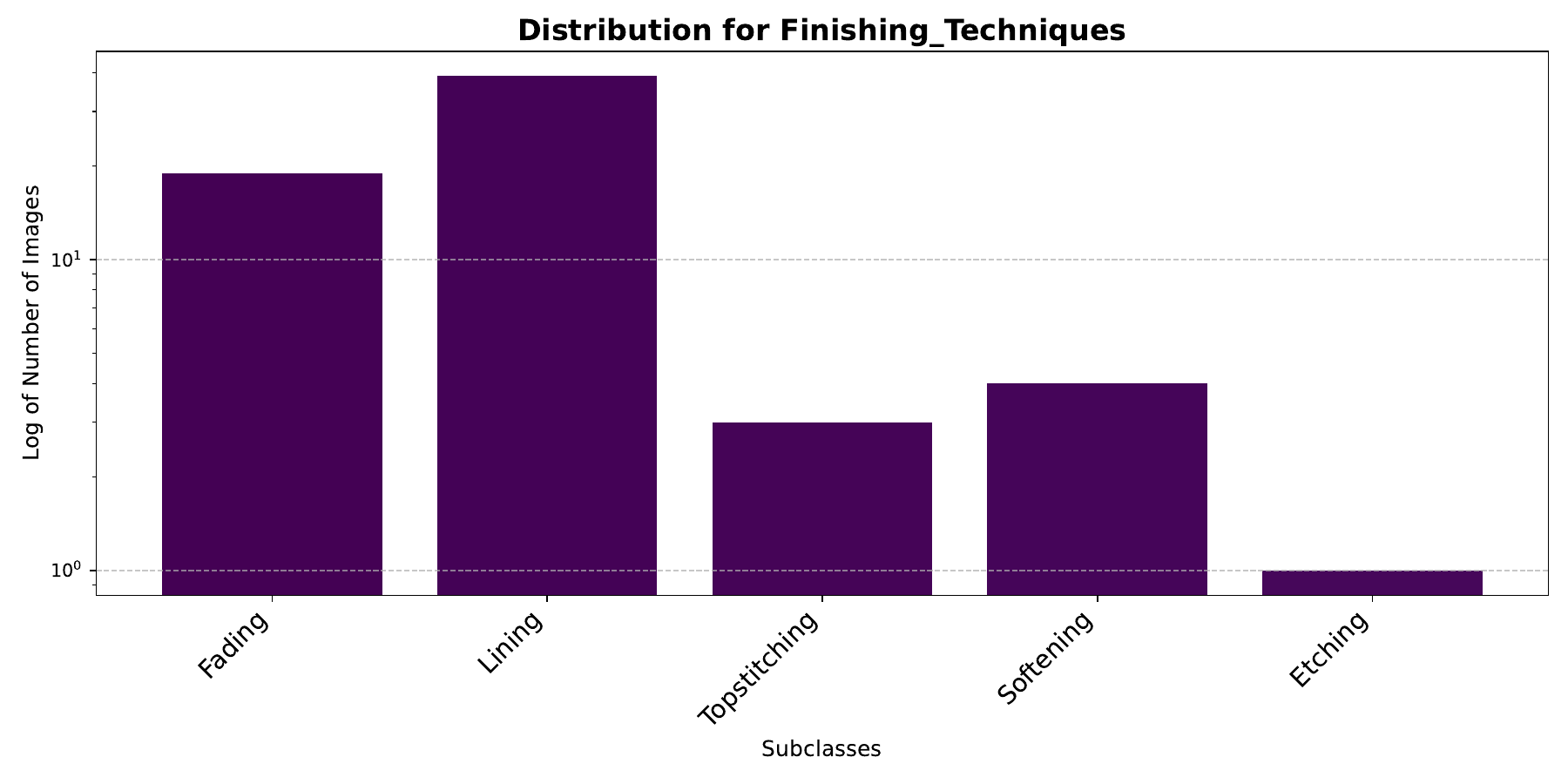}
            \caption{Image Distribution for Finishing techniques}
            \label{fig:lg9}
        \end{figure*}

        Our dataset consists of a wide array of fashion outfit styles, which we categorize into nine distinct classes, each with its own set of sub-classes. These terminologies include \textbf{Garment Types} comprising 29 sub-classes such as \textit{blazers, gowns, hoodies, kimonos, skirts and jackets}. The \textbf{Styles \& Details} class has 17 sub-classes including attributes like \textit{sleeveless, bateau, turtle-neck and halter}. \textbf{Materials \& Patterns} with 21 sub-classes describes fabric types and patterns like \textit{silk, cotton, denim, satin, floral and striped}. \textbf{Construction \& Detailing} includes 40 sub-classes covering elements such as \textit{embroidery, beading, embellishments, metallic, cinched, knitting and vintage or retro aesthetics}. The \textbf{Technical \& Functional} class has five sub-classes which focus on functional features such as \textit{breathability, lightweight materials and detachable components}. \textbf{Occasion \& Aesthetic} includes 18 sub-classes categorizing outfits by their intended use and style such as \textit{casual, formal, traditional, sophisticated, beachwear and whimsical}. \textbf{Construction Techniques} with 10 sub-classes includes garment construction methods like \textit{stitching, buttoning, zippers and lining}. \textbf{Detailing Techniques} with 19 sub-classes describes specific design techniques like \textit{applique, beading, draping, gathering, pleating and layering}. Finally, the \textbf{Finishing Techniques} class has five sub-classes that address finishing touches like \textit{fading, top-stitching and softening}. The distribution of these terminologies across the dataset is presented in log-scaled figures (\ref{fig:lg1} - \ref{fig:lg9}), which provide an insightful representation of their relative prevalence.
        
\section{Fine-Tuning Template for Outfit Sketch Generation}

    We developed a structured prompt template for fine-tuning the models. This template was designed to capture essential aspects of each outfit sketch, including model presence, gender, pose, outfit description, color details and accessories. By standardizing these attributes, we aimed to provide the model with clear, detailed inputs that could guide the generation process toward accurate, high-quality fashion sketches. The template used is as follows:
\begin{verbatim}
MODEL PRESENCE: {MODEL_PRESENCE}
MODEL GENDER: {MODEL_GENDER}
MODEL POSE: {MODEL_POSE}
OUTFIT DESCRIPTION: {OUTFIT_DESCRIPTION}
COLOR DETAILS: {COLOR_DETAILS}
ACCESSORIES: {ACCESSORIES}
\end{verbatim}

\section{Implementation and FLORA Test Set Details}
\textbf{Implementation Details:} All models were fine-tuned with a batch size of 12 using gradient accumulation steps of 12, leveraging the AdamW optimizer for enhanced generalization through weight decay. The initial learning rate was set to \(1 \times 10^{-5}\) and dynamically adjusted with a cosine annealing schedule, 
gradually varying between \(1 \times 10^{-7}\) to \(2 \times 10^{-3}\). 
This scheduling approach allowed for smooth convergence by reducing the learning rate as training progressed. The training was conducted on a robust setup of 4 x 8 Nvidia H100 GPUs with 80GB VRAM.
For all the baselines, elastic weight consolidation \cite{kirkpatrick2017overcoming} was performed to determine where to add LoRA and NeRA.

\textbf{FLORA Test Set:} To demonstrate practical relevance of FLORA, we created a test set of 500 innovative fashion outfit descriptions. Each description was generated using a structured prompt (refer supplementary)  with GPT-4o \cite{gpt4o}, incorporating standard industry-specific terminology. We introduce novel and unique design elements in the prompts to assess the models' capability to interpret both familiar and innovative fashion concepts. Thus, our experiments showcase dataset's effectiveness in real-world applications.

\section{NeRA vs. LoRA: A Mathematical Justification}

This section provides mathematical proof demonstrating why Nonlinear low-rank Expressive Representation Adapters (NeRA) are strictly more expressive than Low-Rank Adaptation (LoRA) adapters. We begin by formally defining LoRA and NeRA, followed by a Taylor series-based proof demonstrating why NeRA provides a superior function approximation. Finally, we connect NeRA’s capabilities with the Kolmogorov–Arnold Representation Theorem to reinforce its universal approximation ability.

\subsection{Recap: LoRA and NeRA Adapter}
\textbf{LoRA - A Linear, Low-Rank Adaptation:} LoRA modifies a model by adding a low-rank update $\Delta W$ to an existing frozen weight matrix $W_0$. Mathematically, this is defined as: 
\begin{equation}
    W = W_0 + \Delta W,
    \quad
    \Delta W = W_{\text{up}} \, W_{\text{down}}
\end{equation}

where:
\begin{itemize}
    \item $W_{\text{down}} \in \mathbb{R}^{d \times r}$ and $W_{\text{up}} \in \mathbb{R}^{r \times d}$
    \item $r \ll d$ ensures the update remains low-rank
\end{itemize}

Given an input $x \in \mathbb{R}^d$, the transformed output is:
\begin{equation}
    \text{LoRA-Output} = W_0 \, x + W_{\text{up}} \,(W_{\text{down}} \, x)
\end{equation}
This transformation remains purely linear and is restricted to rank $r$, meaning LoRA cannot capture nonlinear interactions between features.

\textbf{NeRA Adapter - A More Expressive Nonlinear Adapter:} NeRA adapter builds on LoRA but introduces basis function expansions that allow for nonlinear transformations. The key difference is that NeRA adapter consists of:
\begin{itemize}
    \item A down-projection, which maps input features into a transformed space using nonlinear functions and basis expansions.
    \item An up-projection, which reconstructs an adapted version of the input.
\end{itemize}
Mathematically, this is defined as:
\begin{equation}
H = \phi_{\text{down}}(X)
    \;=\;
    W_b^{\text{down}} \; g(X)
    \;+\;
    W_s^{\text{down}}
    \sum_{i=1}^n c_i \,\gamma_i(X),
\end{equation}
\begin{equation}
X'_{\text{adapter}}
 = \phi_{\text{up}}(H)
    \;=\;
    W_b^{\text{up}} \; g(H)
    \;+\;
    W_s^{\text{up}}
    \sum_{j=1}^m d_j \,\gamma_j(H),
\end{equation}
where:
\begin{itemize}
    \item $g(\cdot)$ is a nonlinear activation function (e.g., ReLU, GeLU),
    \item $\gamma_i(\cdot)$ are basis functions (e.g., Taylor polynomials, B-splines, Fourier expansions),
    \item  $W_s^{\text{down}}$ and $W_s^{\text{up}}$ control the contributions of the basis expansions.
\end{itemize}
Since NeRA adapter explicitly models nonlinear transformations, it can capture complex relationships between features, unlike LoRA.

\subsection{Justification via Taylor Series Expansion}
A key property of NeRA adapter is its ability to approximate arbitrary smooth functions, which LoRA cannot do. We prove this using Taylor series expansions.

\textbf{Step 1: Approximating Functions with Taylor Series}
For any sufficiently smooth function $f: \mathbb{R}^d \to \mathbb{R}^d$ a Taylor series expansion around $x=0$ gives:

\begin{equation*} 
\begin{split}
f_k(x) & = \underbrace{f_k(0)}_{\text{constant term}}
    \;+\;
    \sum_{\alpha} \frac{1}{\alpha!}
    \left(\frac{\partial^{|\alpha|} f_k}{\partial x^\alpha}\Bigl\vert_{x=0}\right) x^\alpha \\
 & + \text{(higher-order remainder)}
\end{split}
\end{equation*}

where,
\begin{itemize}
    \item $\alpha = (\alpha_1, \dots, \alpha_d)$ is a multi-index
    \item $x^\alpha = x_1^{\alpha_1} x_2^{\alpha_2} \cdots x_d^{\alpha_d}$ are monomial terms.
\end{itemize}
Truncating this expansion at order $L$ provides a polynomial approximation:
\begin{equation}
    f(x) \approx P_L(x) = \sum_{i=1}^{N_L} c_i \,\gamma_i(x),
\end{equation}
where, $\gamma_i(x)$ are monomials up to degree $L$.
Since NeRA adapter incorporates these basis functions explicitly, it can approximate any smooth function arbitrarily well—something LoRA cannot achieve with its strictly linear transformations.

\textbf{Step 2: How NeRA adapter Uses Taylor Expansions for Adaptation}

In a NeRA adapter, let the down-projection be:
\begin{equation}
    H \;=\;
    W_b^{\text{down}} \, g(x)
    \;+\;
    W_s^{\text{down}}
    \sum_{i=1}^{N_L} c_i \,\gamma_i(x),
\end{equation}
where,
\begin{itemize}
    \item $\{\gamma_i\}$ are polynomial (Taylor) monomials up to degree $L$.
\end{itemize} 
Then let the up-projection be:
\begin{equation}
    X'_{\text{adapter}}
    \;=\;
    W_b^{\text{up}} \, g(H)
    \;+\;
    W_s^{\text{up}}
    \sum_{j=1}^m d_j \,\gamma_j(H).
\end{equation}
Now $X'_{\text{adapter}}$ is (in principle) a polynomial composition in $x$:
\begin{equation}
    X'_{\text{adapter}}
    \;=\;
    W_b^{\text{up}} \, g\Bigl(W_b^{\text{down}} g(x) + \ldots\Bigr)
    \;+\;
    W_s^{\text{up}} \sum_{j=1}^m d_j \,\gamma_j\Bigl(H\Bigr).
\end{equation}
Expanding $\gamma_j\bigl(H\bigr)$ yields a polynomial in the entries of $H$, each of which is itself a polynomial (plus nonlinearity $g$) in $x$. Provided $g$ is a smooth or well-behaved activation, $X'_{\text{adapter}}$ \emph{can approximate a wide class of continuous functions} $\mathbb{R}^d \to \mathbb{R}^d$, surpassing strictly linear transformations.

\textbf{Step 3: Universality Argument (Local Approximation)}

We now formalize the fact that NeRA adapters can approximate any smooth function locally, while LoRA is limited to low-rank transformations.

\textit{Theorem (Polynomial Approximation):} Let $f: \mathbb{R}^d \to \mathbb{R}^d$ be analytic in a neighborhood of $0$. Then for any $\epsilon > 0$, there exists a polynomial $P_L(x)$ of finite degree $L$ such that,
\begin{equation}
    \|\,f(x) - P_L(x)\| \;<\; \epsilon
\end{equation}
    for all $x$ in some sufficiently small region.

This means that by choosing sufficiently many polynomial terms, NeRA adapter’s basis expansions can approximate any smooth function arbitrarily well within a local region. LoRA, however, is constrained to:
\begin{equation}
    f_{\text{LoRA}}(x) = W_{\text{up}} \,(W_{\text{down}} \, x) + \text{constant terms},
\end{equation}
which is a $rank-r$ linear map in $x$ and cannot capture higher-order (nonlinear) dependencies. This formally proves that NeRA adapter is strictly more expressive than LoRA in local function approximation.

\textbf{Step 4: Kolmogorov–Arnold Representation Theorem Perspective}

While the Taylor series argument above shows \emph{local} universal approximation, the classical Kolmogorov--Arnold Representation Theorem states that \emph{any multivariate continuous function} on a \emph{compact domain} can be written as a sum and composition of \emph{univariate} continuous functions. Formally, in one version:

\textit{Kolmogorov--Arnold Representation:} For any continuous function
\begin{equation}
    F : [0,1]^d \;\to\; \mathbb{R},
\end{equation}
there exist \emph{univariate} continuous functions $\phi_q, \psi_{qk}$ such that
\begin{equation}
    F(x_1, \dots, x_d)
    = 
    \sum_{q=0}^{2d}
    \phi_q\Bigl(
       \sum_{k=1}^d
       \psi_{qk}(x_k)
    \Bigr).
\end{equation}

This theorem implies that if a network architecture can realize \emph{generic compositions of univariate basis expansions}, it can approximate any continuous multivariate function on a cube $[0,1]^d$. NeRA-style adapters, which rely on basis expansions and compositions ($\phi_{\text{down}}$, $\phi_{\text{up}}$), inherently align with this representation principle. Thus, by Kolmogorov-Arnold, NeRA adapter can approximate \emph{any continuous} function, whereas LoRA’s linear approach cannot achieve such universality in higher dimensions.

\subsection{Why NeRA adapter Is Strictly More Expressive than LoRA}

\textbf{LoRA as a Special (Linear) Case of NeRA adapter:} If we \emph{disable} the basis expansions and nonlinearities in NeRA adapter:
\begin{equation}
    W_s^{\text{down}} = 0,\quad
    W_s^{\text{up}} = 0,\quad
    g(x) = x,
\end{equation}
then
\begin{equation}
    \phi_{\text{down}}(x) = W_b^{\text{down}}\,x,
    \quad
    \phi_{\text{up}}(H) = W_b^{\text{up}}\,H,
\end{equation}
giving
\begin{equation}
    X'_{\text{adapter}}
    = W_b^{\text{up}}\,\bigl(W_b^{\text{down}} x\bigr),
\end{equation}
which is precisely a rank-$n$ linear map akin to LoRA. Thus, \emph{LoRA is embedded in NeRA adapter} as a degenerate (linear-only) configuration.

\textbf{Nonlinear and Higher-Order Terms in NeRA adapter:} In general, NeRA adapter includes expansions:
\begin{equation}
    \sum_{i} c_i \gamma_i(x),
    \quad
    \gamma_i(x) = \text{(e.g.\ polynomial monomial)}.
\end{equation}
Hence, $X'_{\text{adapter}}$ can contain \emph{terms of the form} $x_1^{\alpha_1} x_2^{\alpha_2} \cdots x_d^{\alpha_d}$ or other advanced expansions (Fourier, spline, etc.). \emph{No purely linear LoRA block can replicate such higher-order interactions} without artificially stacking multiple linear LoRA blocks across many layers.

\subsection{Proof Summary}
\begin{itemize}
    \item \textbf{Taylor-series-based argument}: 
    By leveraging polynomial expansions up to order $L$, NeRA can approximate local behavior of smooth functions to arbitrary precision. LoRA, being linear and low-rank, lacks this capacity.
    \item \textbf{Kolmogorov--Arnold-based argument}:
    NeRA’s composition of univariate basis expansions aligns with the Kolmogorov-Arnold Theorem, thus enabling universal approximation for continuous functions on compact domains in $\mathbb{R}^d$.
    \item \textbf{LoRA is a special case}: 
    Setting all basis expansions to zero and removing nonlinearities recovers a linear rank-$r$ form, demonstrating that NeRA \emph{strictly generalizes} LoRA.
\end{itemize}
\bigskip
Therefore, \textbf{NeRA can realize richer, higher-order function approximations}, surpassing the purely linear rank-$r$ approach of LoRA. Consequently, \textbf{NeRA adapter works better than LoRA} in scenarios demanding expressive, nonlinear transformations while retaining the efficient, adapter-based paradigm.

In practice, a NeRA may incorporate polynomial or spline expansions of a modest order, often capturing crucial nonlinearities with fewer parameters than a naive full-rank update. Empirical results generally confirm that NeRA-like architectures can adapt large models more flexibly than LoRA alone, especially under distribution shifts or tasks requiring higher-order feature interactions.

\end{document}